\documentclass[sigconf,screen,balance]{acmart}
\AtBeginDocument{%
  \providecommand\BibTeX{{%
    \normalfont B\kern-0.5em{\scshape i\kern-0.25em b}\kern-0.8em\TeX}}}

\setcopyright{acmcopyright}
\copyrightyear{2021}
\acmYear{2021}
\acmDOI{10.1145/1122445.1122456}

\acmConference[MM '21]{MM '21: ACM International Conference on Multimedia}{October 20--24, 2021}{Chengdu, China}
\acmBooktitle{MM '18: ACM International Conference on Multimedia,
  October 20--24, 2021, Chengdu, China}
\acmPrice{15.00}
\acmISBN{978-1-4503-XXXX-X/18/06}

\usepackage{wrapfig}

\usepackage{hyperref}       
\usepackage{url}            
\usepackage{booktabs}       
\usepackage{amsfonts}       
\usepackage{nicefrac}       
\usepackage{microtype}
\usepackage{multirow}

\usepackage{caption}
\usepackage{enumitem}
\setitemize{noitemsep,topsep=0pt,parsep=0pt,partopsep=0pt}

\acmSubmissionID{462}

\begin{document}

\title{Cross-View Exocentric to Egocentric Video Synthesis}


 \author{Gaowen Liu}
 \email{gaoliu@cisco.com}
 \affiliation{%
   \institution{Cisco Systems}
   \city{San Jose, CA}
   \country{USA}
 }
 \author{Hao Tang}
 \email{hao.tang@unitn.it}
 \affiliation{%
   \institution{DISI, University of Trento}
   \city{Trento}
   \country{Italy}}

 \author{Hugo Latapie}
  \email{hlatapie@cisco.com}
 \affiliation{%
   \institution{Cisco Systems}
   \city{San Jose, CA}
   \country{USA}
 }

 \author{Jason Corso}
 \email{jcorso@stevens.edu}
 \affiliation{%
  \institution{Stevens Institute of Technology}
  \city{Hoboken, NJ}
  \country{USA}}

 \author{Yan Yan}
 \email{yyan34@iit.edu}
 \affiliation{%
   \institution{Illinois Institute of Technology}
   \city{Chicago, IL}
   \country{USA}}




\renewcommand{\shortauthors}{Trovato and Tobin, et al.}

\begin{abstract}
Cross-view video synthesis task seeks to generate video sequences of one view from another dramatically different view. In this paper, we investigate the exocentric (third-person) view to egocentric (first-person) view video generation task. This is  challenging because egocentric view sometimes is remarkably different from the exocentric view. Thus, transforming the appearances across the two different views is a non-trivial task. Particularly, we propose a novel Bi-directional Spatial Temporal Attention Fusion Generative Adversarial Network (STA-GAN) to learn both spatial and temporal information to generate egocentric video sequences from the exocentric view. The proposed STA-GAN consists of three parts: temporal branch, spatial branch, and attention fusion. First, the temporal and spatial branches generate a sequence of fake frames and their corresponding features. The fake frames are generated in both downstream and upstream directions for both temporal and spatial branches. Next, the generated four different fake frames and their corresponding features (spatial and temporal branches in two directions) are fed into a novel multi-generation attention fusion module to produce the final video sequence. Meanwhile, we also propose a novel temporal and spatial dual-discriminator for more robust network optimization. Extensive experiments on the Side2Ego and Top2Ego datasets~\cite{third2019} show that the proposed STA-GAN significantly outperforms the existing methods.
\end{abstract}
\ccsdesc[500]{Computer systems organization~Embedded systems}
\ccsdesc[300]{Computer systems organization~Redundancy}
\ccsdesc{Computer systems organization~Robotics}
\ccsdesc[100]{Networks~Network reliability}
\keywords{Cross-view Video Synthesis, Exocentric, Egocentric}

\maketitle

\section{Introduction}

\begin{figure}[t]
\centering
\includegraphics[width=1\linewidth]{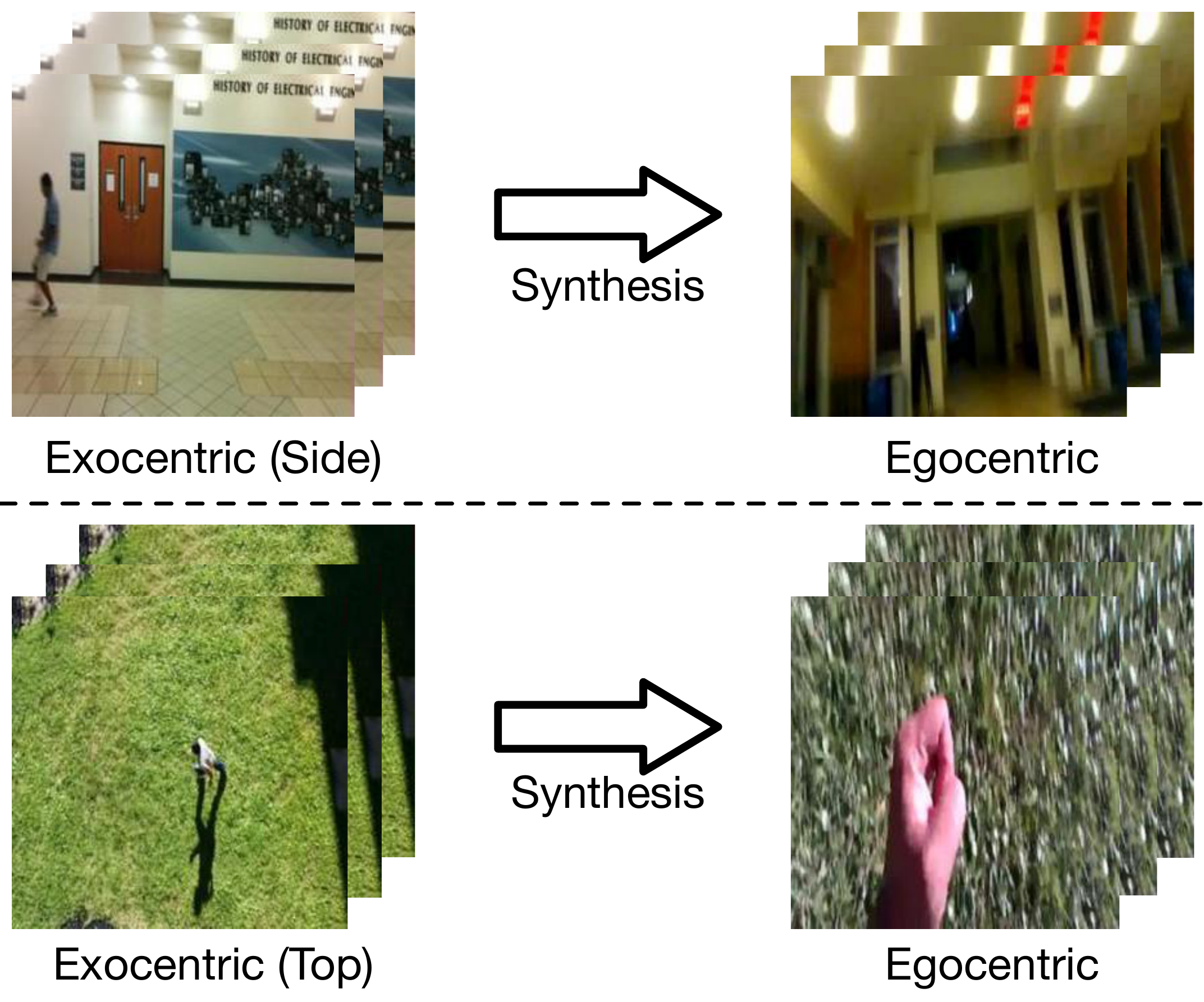}
\caption{The goal of exocentric to egocentric cross-view video synthesis is to generate video sequence from exocentric perspective (Side/Top) to egocentric perspective.}
	\vspace{-0.2cm}
\label{fig:goal}
\end{figure}

Wearable cameras, also known as first-person cameras, are widely used in our daily lives since the appearance of low price but high-quality wearable products such as Google Clips, GoPro cameras~\cite{Gopro}. Meanwhile, egocentric (first-person) vision has become a critical research topic in the computer vision field ~\cite{yanego_2015,ego2top,firstperson_kanade,third2019,liu2020exocentric}. As we know, first-person egocentric views have some unique properties other than third-person exocentric views. Traditional exocentric cameras usually give a wide and global view of the high-level appearance in a video. However, egocentric cameras are able to reveal the focus of attention, behavior, and goal of its wearer. Early egocentric vision studies~\cite{firstperson_kanade} found that humans are able to seamlessly transfer knowledge between egocentric and exocentric perspectives when performing different activities or interacting with objects.
Therefore, understanding the relationship between egocentric and exocentric views is a critical need in computer vision. 

\begin{figure*}[t]
	\centering
	\includegraphics[width=1\linewidth]{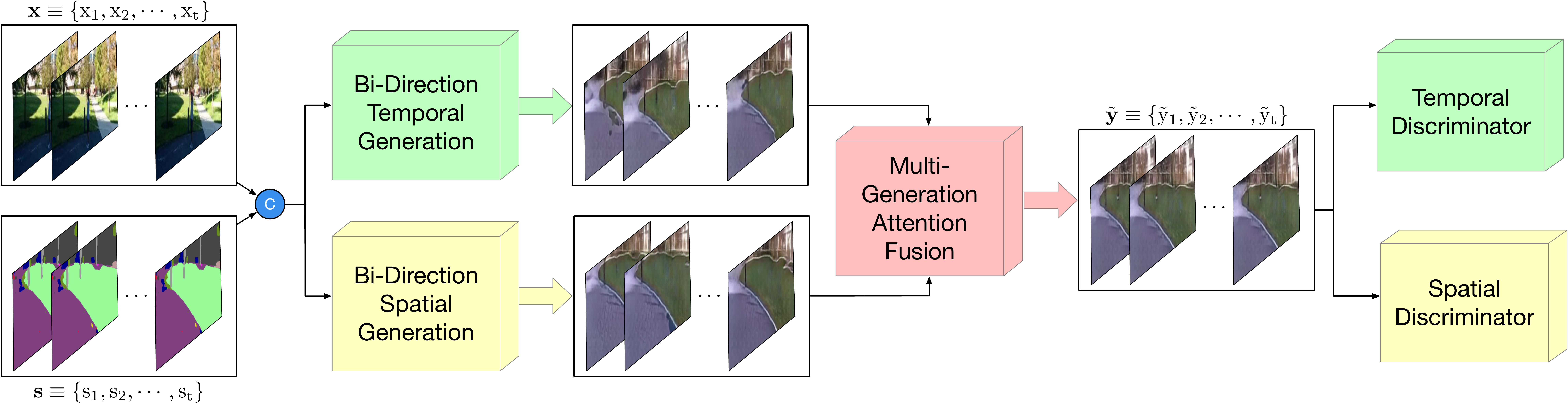}
	\caption{The framework of the proposed STA-GAN, which consists of four parts, i.e.,
		a temporal generation branch, a spatial generation branch, an attention fusion module and a temporal and spatial dual-discriminator. 
		The bi-directional spatial and temporal generation branches accept exocentric video sequence and conditional semantic maps as inputs and simultaneously synthesizes egocentric video sequence. Then multi-generation attention fusion module fuses the synthesized video sequences that obtained from temporal and spatial generation branches and outputs the final egocentric view video sequence.
		The proposed dual-discriminator aims to distinguish the generated videos from two spaces, i.e., temporal space and spatial space.}
	\vspace{-0.2cm}
	\label{fig:framework}
\end{figure*}

However, there is little research to address this important problem in the literature.  One likely reason is the difficulty in collecting a large amount of high quality egocentric view data for the wide variety of computer vision problems in different emerging context~\cite{Poleg14,yanego_2015}. To address such a limitation, the technique of generating egocentric videos from exocentric videos, as shown in~Figure~\ref{fig:goal}, becomes an alternative solution. Nevertheless, generating egocentric data is extremely challenging because significant differences of visual appearance are expected between egocentric and exocentric videos. Moreover, the sharp changes in the viewpoint makes it an even more difficult task. 

Based on these observations, in this paper, we propose a Generative Adversarial Network (GAN) based video generation approach to bridge egocentric and exocentric video analysis. In the past few years, GANs~\cite{NIPS2014_5423} are extremely successful in various image generation problems~\cite{isola2017image,duan2021cascade,Regmi_2018_CVPR,zhang2018self,brock2018large,karras2019style,shama2019adversarial,huh2019feedback,tang2020local}. Particularly, X-Fork~\cite{Regmi_2018_CVPR}, X-Seq~\cite{Regmi_2018_CVPR} and SelectionGAN~\cite{tang2019multichannel} are proposed to tackle cross-view image generation tasks. However, these approaches do not aim at more challenging cross-view video generation tasks. GANs are also developed for video generation in literature. RecycleGAN \cite{Recycle-GAN} works on unsupervised video retargeting for various domains such as face retargeting. Wang et al.~\cite{wang2018vid2vid} propose vid2vid framework which is able to transform a sequence of semantic representations, e.g., semantic label map and sketch map, to a sequence of video frames. StarGAN \cite{choi2019stargan} allows simultaneous training of multiple datasets with different domains within a single network. However, these methods are not able to generate satisfactory results on cross-view video generation tasks due to the dramatically differences between exocentric and egocentric views. 

Overcoming these limitations, we propose a novel Bi-directional Spatial Temporal Attention fusion GAN (STA-GAN) to generate egocentric video from an exocentric perspective. 
The proposed framework consists of three parts: temporal generation branch, spatial generation branch, and attention fusion module, as shown in Figure~\ref{fig:framework}. First, both temporal and spatial branches take a sequence of exocentric view frames as the input and generate a sequence of fake frames of egocentric view and their feature maps for each module, respectively. The fake frames and the corresponding feature maps are generated simultaneously in both downstream and upstream directions. Therefore, each input frame are corresponding to four pairs of fake frames and feature maps (spatial and temporal modules in two directions). Second, the generated four pairs of fake frames and feature maps are fed into the attention fusion module. 
Finally, we generate the fused output of fake frames. 
The proposed framework is able to learn both spatial and temporal information from the forward and backward  directions in the time domain simultaneously. We demonstrate that the proposed STA-GAN outperforms other baselines such as X-Fork~\cite{Regmi_2018_CVPR}, X-Seq~\cite{Regmi_2018_CVPR}, SelectionGAN~\cite{tang2019multichannel}, RecycleGAN~\cite{Recycle-GAN}, vid2vid~\cite{wang2018vid2vid} through extensive experimental evaluations. We establish state-of-the-art results on the Side2Ego and Top2Ego datasets~\cite{third2019}. To the best of our knowledge, we are the first to attempt to incorporate a bi-directional spatial temporal  generative network for exocentric to egocentric cross-view video synthesis.


The contributions of this paper can be summarized as follows:

\begin{itemize}
\item A novel Bi-directional Spatial Temporal Attention Fusion Generative Adversarial Network (STA-GAN) is proposed. It aims at deploying temporal and spatial information in video and learning both spatial and temporal information between different views simultaneously.   
\item A group of novel downstream/upstream temporal and spatial loss functions are designed for neural network training. Moreover, a novel attention fusion module is proposed to fuse the generated fake frames to obtain refined final results. Meanwhile, a novel  temporal and spatial dual-discriminator are proposed for network training.
\item Experimental results on different cross-view datasets show the effectiveness of the proposed model. Our approach outperforms state-of-the-art results by a large margin for the cross-view exocentric to egocentric  video synthesis. To the best of our knowledge, we are the first to attempt to tackle the cross-view exocentric to egocentric video generation task. 
\end{itemize}
%

\section{Related Work}
\noindent \textbf{Generative Adversarial Networks (GANs).} Over the last few years, GANs~\cite{NIPS2014_5423} have been shown effectively in many image generation and translation tasks~\cite{isola2017image,CycleGAN2017,tang2020unified,shaham2019singan,liu2019few,tang2020xinggan,tang2020dual,tang2020bipartite,tang2019cycle,tang2018gesturegan}.
For example, Isola et al.~\cite{isola2017image} propose Pix2Pix adversarial learning framework for paired image generation. Zhu et al.~\cite{CycleGAN2017} introduce CycleGAN which developed cycle-consistency constraint to deal with unpaired image generation. However, these works aim to generate images which have a large degree of overlapping in the appearance and view with input images. 
Synthesis is much more challenging when the generation is conditioned on images with drastically different views. Recently, researchers investigate cross-view image generation problems~\cite{REGMI2019}. This is a more challenging task since different views share little overlap information. To tackle this problem, Krishna et al.~\cite{Regmi_2018_CVPR} propose X-Fork and X-Seq GAN-based architecture using an extra semantic map to facilitate generation. Tang et al.~\cite{tang2019multichannel} propose a semantic-guided multi-channel attention selection module within a GAN framework for cross-view image generation. However, these methods are limited to cross-view image  generation task, they are not able to generate satisfactory results for cross-view video generation. 

\noindent \textbf{Egocentric Vision.} 
Egocentric vision has been recently explored in the computer vision field~\cite{omid,FathiCVPR2012,Pirsi,ogaki2012coupling,taralova2011source,FathiECCV2012,Fathi11,Poleg14,Yonetani16}. Aghazadeh et al.~\cite{omid} propose an approach for discovering anomalous events from videos captured from a small camera attached to a person's chest. Fathi et al.~\cite{FathiCVPR2012} introduce a method for individuating social interactions in first-person videos collected during social events. Some recent works~\cite{Pirsi,ogaki2012coupling,taralova2011source,Fathi11,Ryoo13} have focused on activity analysis considering different scenarios (e.g., kitchen, office, home). Xu et al.~\cite{Xu17} propose a semi-Siamese CNN architecture to address the person-level correspondences across first- and third-person videos. They formulate the problem as learning a joint embedding space for first- and third-person videos that considers both spatial- and motion-domain cues.

\noindent \textbf{Video-to-Video Synthesis.} There is few recent work investigate video generation problem~\cite{TGAN2017,FathiECCV2012,chan2019dance,stovid}. TGAN~\cite{TGAN2017} directly generate video clips from noise by using generative adversarial networks. MoCoGAN~\cite{MoCoGAN} employ unsupervised adversarial training to decompose motion and content to control the image-to-video generation. Pan et al.~\cite{seg2vid_pan} work on video-to-video translation to generate a sequence of frames from a sequence of aligned semantic representations. Some recent works such as RecycleGAN~\cite{Recycle-GAN} and Vid2Vid~\cite{wang2018vid2vid} learn mapping between different videos and transferred motion between faces and from poses to body, respectively. Frameworks~\cite{TDB16a,couplegan,zhou2016view,tvsn_cvpr2017} propose image generation networks for 3D view synthesis. 

However, existing frameworks on video generation require the input and output video scenes sharing the similar architecture, which were insufficient for cross-view video generation. Particularly, exocentric to egocentric cross-view video generation has not yet been studied in literature yet. Our method investigates both cross-view generation and video generation in the exocentric to egocentric perspective setting, which is more challenging than various video generation problems. To the best of our knowledge, this is the first attempt in literature.  
\section{Bi-directional Spatial Temporal Attention Fusion GANs}
In this section, we present the details of the proposed Bi-directional Spatial Temporal Attention Fusion GAN (STA-GAN). The overall framework of the proposed STA-GAN is illustrated in Figure~\ref{fig:framework}, which contains four different modules, i.e., temporal generation module, spatial generation module, attention fusion, and a dual-discriminator. The bi-directional temporal generation module learns the temporal information of the target video. Meanwhile, the bi-directional spatial generation module models the spatial information of video frames. Moreover, the multi-generation attention fusion module fuses the information from temporal and spatial modules. 
Lastly, the proposed temporal and spatial dual discriminator aims to distinguish the generated videos from two spaces, i.e., temporal space and spatial space.

\sloppy
\subsection{Semantic-guided Cross-view Video Generation} 
Our goal is to generate a video sequence of egocentric view $\rm \textbf{y} {\equiv} {\{y_1,...,y_t\}}$ from a video sequence of exocentric view $\rm \textbf{x} {\equiv} {\{x_1,...,x_t\}}$, where $\rm y_t$ and $\rm x_t$ are corresponding real video frames. Our task is to learn a video generator $G$ receives  $ \textbf{x}$ and output $\rm \tilde{\textbf{y}}$ close to the real video $\rm {\textbf{y}}$. This process can be formulated as:
\begin{equation}
\begin{aligned}
{ \tilde{\textbf{y}}} = G( \textbf{x}).
\end{aligned}
\label{eqn:generat}
\end{equation}
However, the cross-view exocentric to egocentric video synthesis task is challenging due to several reasons. First, exocentric and egocentric views have little overlapping, which leads to ambiguity issues in the generation process. Second, the existing egocentric view datasets are rare and collected by wearable devices which leads huge amount of blurry videos.
To alleviate both limitations, in this work, we employ a semantic-guided strategy. We incorporate semantic maps as a conditional guidance.  
Specifically, we adopt RefineNet~\cite{Lin:2017:RefineNet,lin2019refinenet,tang2019multichannel} to generate semantic maps on both Side2Ego and Top2Ego datasets~\cite{third2019}. 
The generated semantic maps are used as the conditional input of the generator $G$, as shown in Figure~\ref{fig:framework}.
We concatenate the input video $\rm \textbf{x}$ from the exocentric view and the semantic map $\rm \textbf{s}{\equiv}{\{s_1, ..., s_t\}}$ from a egocentric view, and input them into the video generator $G$ and synthesize the egocentric view video sequence $\rm \tilde{\textbf{y}}$ as:
\begin{equation}
\begin{aligned}
{\rm \tilde{\textbf{y}} {=}} G(\rm \textbf{x}, \textbf{s}).
\end{aligned}
\label{eqn:generat}
\end{equation}
In this way, the semantic maps provide stronger supervision to guide the cross-view video synthesis.
\begin{figure*}[t]
\center
{\includegraphics[width=0.49\linewidth]{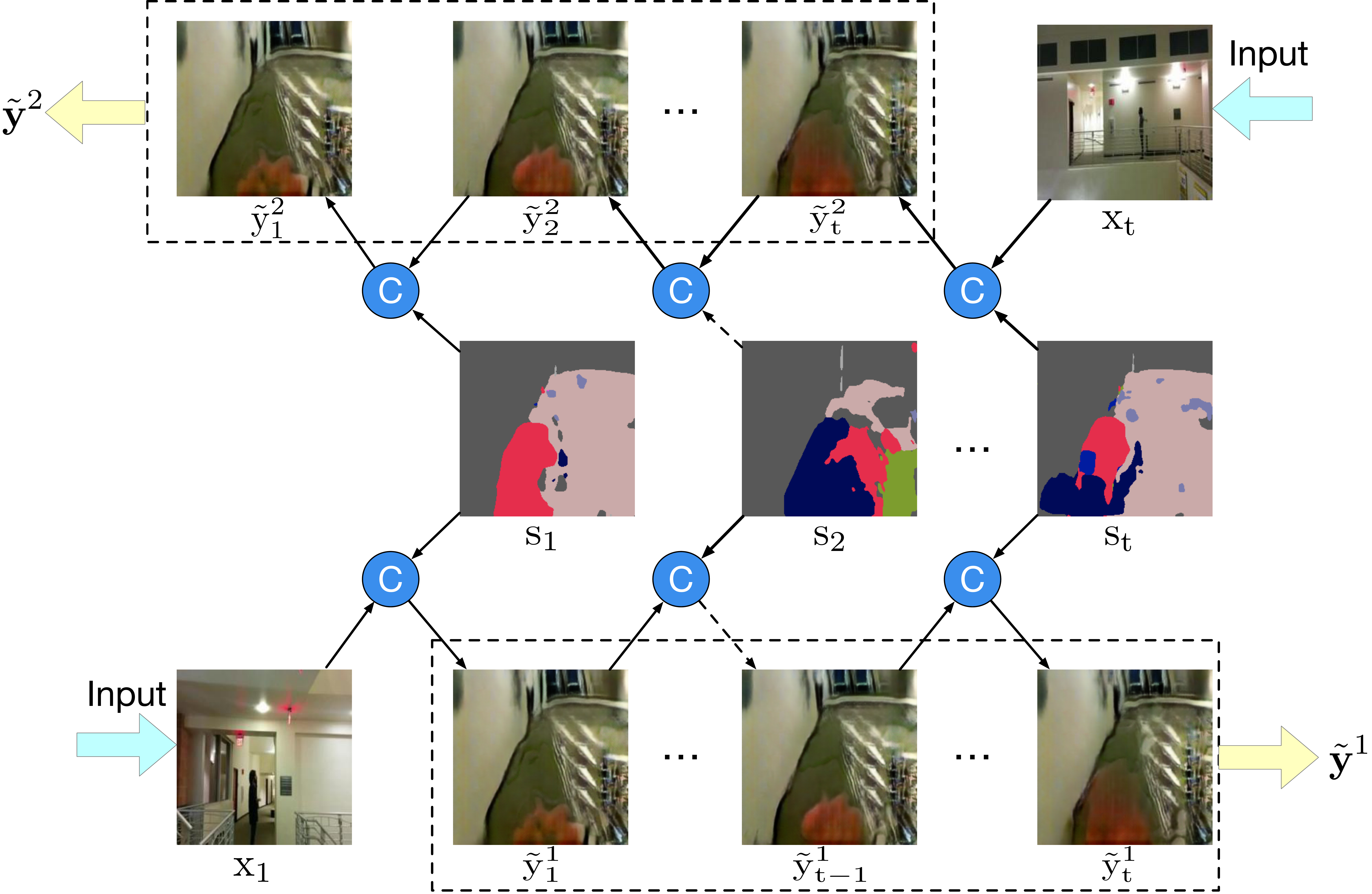}} 
{\includegraphics[width=0.49\linewidth]{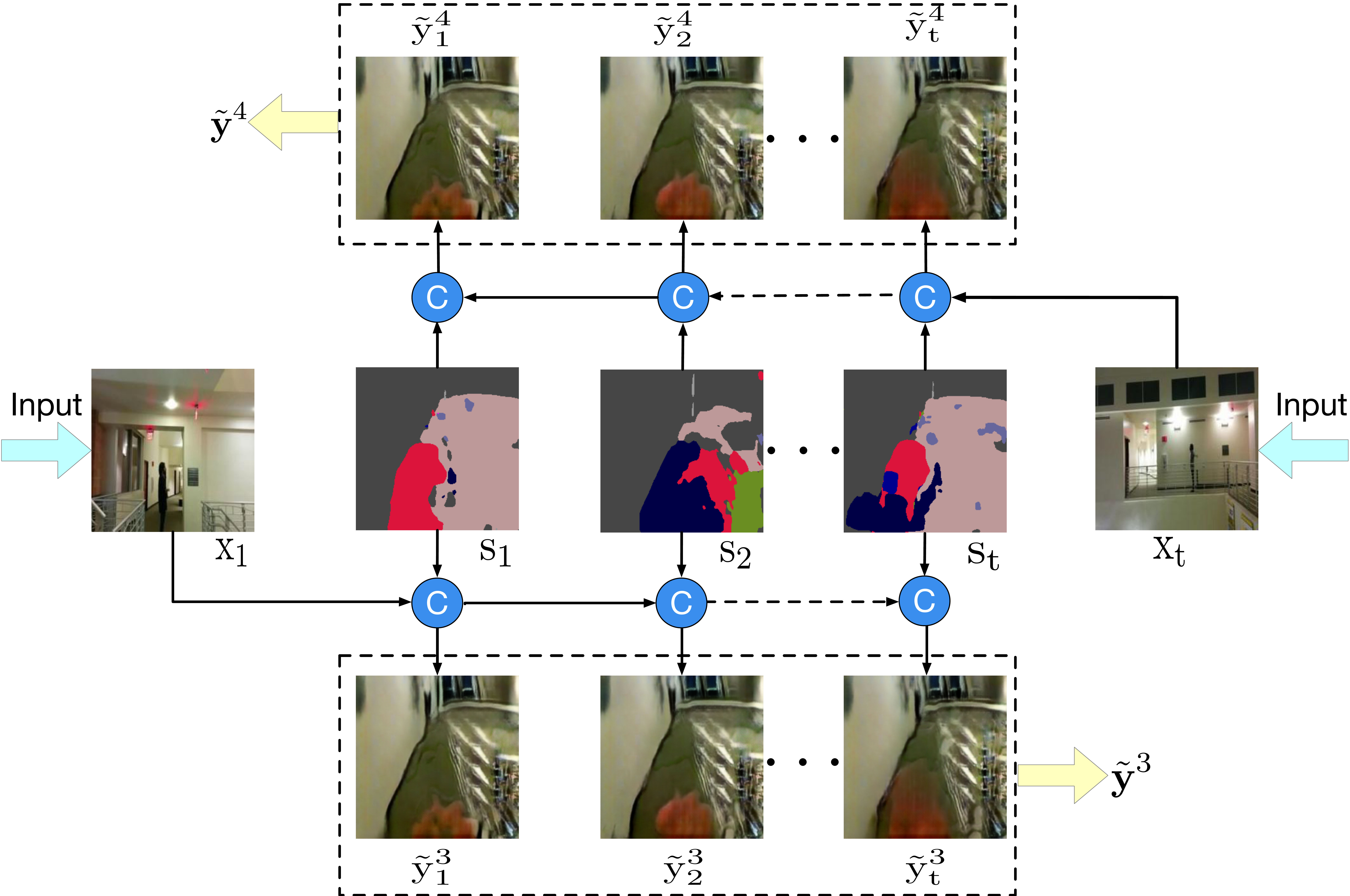}} 
\caption{Illustration of the proposed \textbf{(left)} temporal module, and \textbf{(right)} spatial module. The symbol \textcircled{c} denotes concatenation. This channel-wise concatenation operating is able to avoid mixing of different modalities of information (RGB images and semantic maps). The concatenation will be used in the later attention fusion module.}
	\vspace{-0.2cm}
\label{fig:temporal}
\end{figure*}
\subsection{Bi-directional Temporal Generation} 
Due to the complexity and particularity of video generation task, we take into account of time information in video sequences. Temporal information is crucial in video analysis, a video frame is usually highly correlated to adjacent frames. 
In this work, we enforce a temporal coherence between adjacent frames by integrating adjacent semantic maps as guidance input in both downstream and upstream directions, which are shown along the arrows in Figure~\ref{fig:temporal}(left). 
The conditional semantic map $\rm s_1$ together with the input frame $\rm x_1$ are input into the generator $G$, and produce the synthesized frame ${\rm \tilde{y}_1^1} {=}G \rm (x_1, s_1)$. Then generated ${\rm \tilde{y}_1^1}$ and the next semantic map $\rm s_2$ are further fed into the generator $G$ which reconstructs a new egocentric frame ${\rm \tilde{y}_2^1}$. We generate downstream direction of egocentric frame at time $\rm t$ based on $\rm \tilde{y}_{t-1}^1$ and semantic map $\rm s_t$. We can formalize the downstream process as: 
\begin{equation}
\begin{aligned}
{\rm \tilde{y}_t^1} &=  G \rm ( \tilde{y}_{t-1}^1, s_t).
\end{aligned}
\label{eqn:spatial_down}
\end{equation}
In this way, we are able to obtain the generated video sequence $\rm \tilde{\textbf{\textbf{y}}}^1 {\equiv} {\{\tilde{y}_1^1,...,\tilde{y}_t^1\}}$.

In the opposite direction, the upstream process is time reverse generation process and starts from time $\rm t$. 
We input $\rm x_t$ and $\rm s_t$ to the generator $G$ to produce the first frame ${\rm \tilde{y}_t^2}  = G \rm (x_t, s_{t})$. 
Then other frames can be obtained by using the following formulation:
\begin{equation}
\begin{aligned}
{\rm \tilde{y}_{t-1}^2} &=G\rm ( \tilde{y}_{t}^2, s_{t-1}).
\end{aligned}
\label{eqn:spatial_down}
\end{equation}
By doing so, we are able to obtain the generated video sequence $\rm \tilde{\textbf{\textbf{y}}}^2 {\equiv} {\{\tilde{y}_1^2,...,\tilde{y}_t^2\}}$.

To better model the temporal information, we further propose a novel temporal loss.
The temporal loss composes of both downstream process and upstream process. 
The key idea is to add temporal constraints to the generation process. 
For each frame, the downstream temporal loss is:
\begin{equation}
\begin{aligned}
\mathcal{T}_{dn}(\rm \tilde y_t^{1}) &= \mathbb{E}_{\rm x_{t}, \rm \tilde y_{t}^{1}} \left[ \|\rm y_{t}-\rm \tilde y_{t}^{1}\|_1 \right] + \mathbb{E}_{\rm x_t, \rm y_t} \left[ \log D_S(\rm x_t, \rm y_t) \right]\\
& + \mathbb{E}_{\rm x_t, \rm \tilde y_t^{1}} \left[\log (1 - D_S({\rm x_t}, \rm \tilde{y}_t^1)) \right]. 
\end{aligned}
\label{eqn:timedn}
\end{equation}
Overall, the downstream temporal loss can be represented as:
\begin{equation}
\begin{aligned}
\mathcal{T}_{dn}(\rm \textbf{x},\rm \tilde{\textbf{y}}^1) = 
&\sum_{t=1}^{t} \lambda_u\mathcal{T}_{dn}(\rm \tilde y_t^{1}).
\end{aligned}
\label{eqn:timedl}
\end{equation}
The upstream temporal loss is formulated similar as downstream temporal loss, however they computed in an opposite direction for each frame as:
\begin{equation}
\begin{aligned}
\mathcal{T}_{up}(\rm \tilde y_t^{2}) &= \mathbb{E}_{\rm x_t, \rm \tilde y_{t}^{2}} \left[ \|\rm y_{t}-\rm \tilde y_{t}^{2}\|_1 \right] + \mathbb{E}_{\rm x_t, \rm y_t} \left[ \log D_S(\rm x_t, \rm y_t) \right]\\ 
& + \mathbb{E}_{\rm x_t, \rm \tilde y_t^{2}} \left[\log (1 - D_S({\rm x_t}, \rm \tilde{y}_t^2)) \right].
\end{aligned}
\label{eqn:timedn}
\end{equation}
The overall upstream temporal loss is:
\begin{equation}
\begin{aligned}
\mathcal{T}_{up}(\rm \textbf{x},\rm \tilde{\textbf{y}}^2) = 
&\sum_{t=t}^{1} \lambda_d\mathcal{T}_{up}(\rm \tilde y_t^{2}).
\end{aligned}
\label{eqn:timeul}
\end{equation}
Finally, the temporal loss is the sum of Equation~\eqref{eqn:timedl} and Equation~\eqref{eqn:timeul},
\begin{equation}
\begin{aligned}
\mathcal{L}_{T}(\rm \textbf{x},\rm \tilde{\textbf{y}}) = 
&\mathcal{T}_{dn}(\rm \textbf{x},\rm \tilde{\textbf{y}}^1) + \mathcal{T}_{up}(\rm \textbf{x},\rm \tilde{\textbf{y}}^2).
\end{aligned}
\label{eqn:temporalloss}
\end{equation}


\subsection{Bi-directional Spatial Generation} 
Spatial information plays a crucial role in various video related tasks, such as activity recognition~\cite{CVPR2019_CycleTime}, object recognition~\cite{STLattice2018CVPR}, etc. In this work, we incorporate the effects of spatial information by generating non-adjacent frames using the corresponding semantic maps. As illustrated in the spatial module of Figure~\ref{fig:temporal}(right), the process of downstream direction is generated along the arrows:
\begin{equation}
\begin{aligned}
{\rm \tilde y_1^{3} =} G{\rm (x_1, s_1)}, \cdots, {\rm \tilde y_t^{3} =} G \rm (x_1, s_t).
\end{aligned}
\label{eqn:spatial_down}
\end{equation}
In this way, we are able to obtain the generated video sequence $\rm \tilde{\textbf{\textbf{y}}}^3 {\equiv} {\{\tilde{y}_1^3,...,\tilde{y}_t^3\}}$.
The opposite upstream direction in which the sequences is formulated as:
\begin{equation}
\begin{aligned}
{\rm \tilde y_t^{4} =} G{\rm (x_t, s_t), \cdots, \tilde y_1^{4} =} G \rm (x_t, s_1),
\end{aligned}
\label{eqn:spatial_up}
\end{equation}
where we are able to obtain the generated video sequence $\rm \tilde{\textbf{\textbf{y}}}^4 {\equiv} {\{\tilde{y}_1^4,...,\tilde{y}_t^4\}}$.

To learn the spatial information better, we propose a new spatial loss.
The spatial loss composes of two parts, which are downstream spatial loss and upstream spatial loss. The intuition is that activities and events in videos are spatially related between adjacent frames and are reversible. 
For each frame, the downstream spatial loss is formulated as follows:
\begin{equation}
\begin{aligned}
\mathcal{S}_{dn}(\rm \tilde y_t^{3}) &= \mathbb{E}_{\rm x_{t-i},\rm \tilde y_{t}^{3}} \left[ \|\rm y_{t}-\rm \tilde y_{t}^{3}\|_1 \right] + \mathbb{E}_{\rm x_{t-i}, \rm y_t} \left[ \log D_S(\rm x_{t-i}, \rm y_t) \right] \\ 
& + \mathbb{E}_{\rm x_{t-i}, \rm \tilde y_t^{3}} \left[\log (1 - D_S(\rm x_{t-i}, \rm \tilde y_t^{3})) \right].
\end{aligned}
\label{eqn:sdn}
\end{equation}
The overall downstream spatial loss is:
\begin{equation}
\begin{aligned}
\mathcal{S}_{dn}(\rm \textbf{x},\rm \tilde{\textbf{y}}^3) = 
&\sum_{t=1}^{t} \lambda_n\mathcal{S}_{dn}(\rm \tilde y_t^{3}).
\end{aligned}
\label{eqn:sdloss}
\end{equation}
The upstream spatial loss is generated reversely, which is formulated similar as: 
\begin{equation}
\begin{aligned}
\mathcal{S}_{up}(\rm \tilde y_t^{4}) &= \mathbb{E}_{\rm x_{t+i}, \rm \tilde y_{t}^{4}} \left[ \|\rm y_{t}-\rm \tilde y_{t}^{4}\|_1 \right] + \mathbb{E}_{\rm x_{t+i}, \rm y_t} \left[ \log D_S(\rm x_{t+i}, \rm y_t) \right]\\
& + \mathbb{E}_{\rm x_{t+i}, \rm \tilde y_t^{4}} \left[\log (1 - D_S(\rm x_{t+i}, \rm \tilde y_t^{4})) \right].
\end{aligned}
\label{eqn:sup}
\end{equation}
The overall upstream spatial loss:
\begin{equation}
\begin{aligned}
\mathcal{S}_{up}(\rm \textbf{x},\rm \tilde{\textbf{y}}^4) = 
&\sum_{t=t}^{1} \lambda_p\mathcal{T}_{up}(\rm \tilde y_t^{4}).
\end{aligned}
\label{eqn:suploss}
\vspace{-0.1cm}
\end{equation}
The overall spatial loss is the sum of Equation~(\ref{eqn:sdloss}) and Equation~(\ref{eqn:suploss}), where $i$ is the time truncate during training:
\begin{equation}
\begin{aligned}
\mathcal{L}_{S}(\rm \textbf{x},\rm \tilde{\textbf{y}}) = 
&\mathcal{S}_{dn}(\rm \textbf{x},\rm \tilde{\textbf{y}}^3) + \mathcal{S}_{up}(\rm \textbf{x},\rm \tilde{\textbf{y}}^4).
\end{aligned}
\label{eqn:temporalloss}
\end{equation}

\subsection{Multi-generation Attention Fusion} 
\begin{figure}[tbp]
\centering
{\includegraphics[width=1\linewidth]{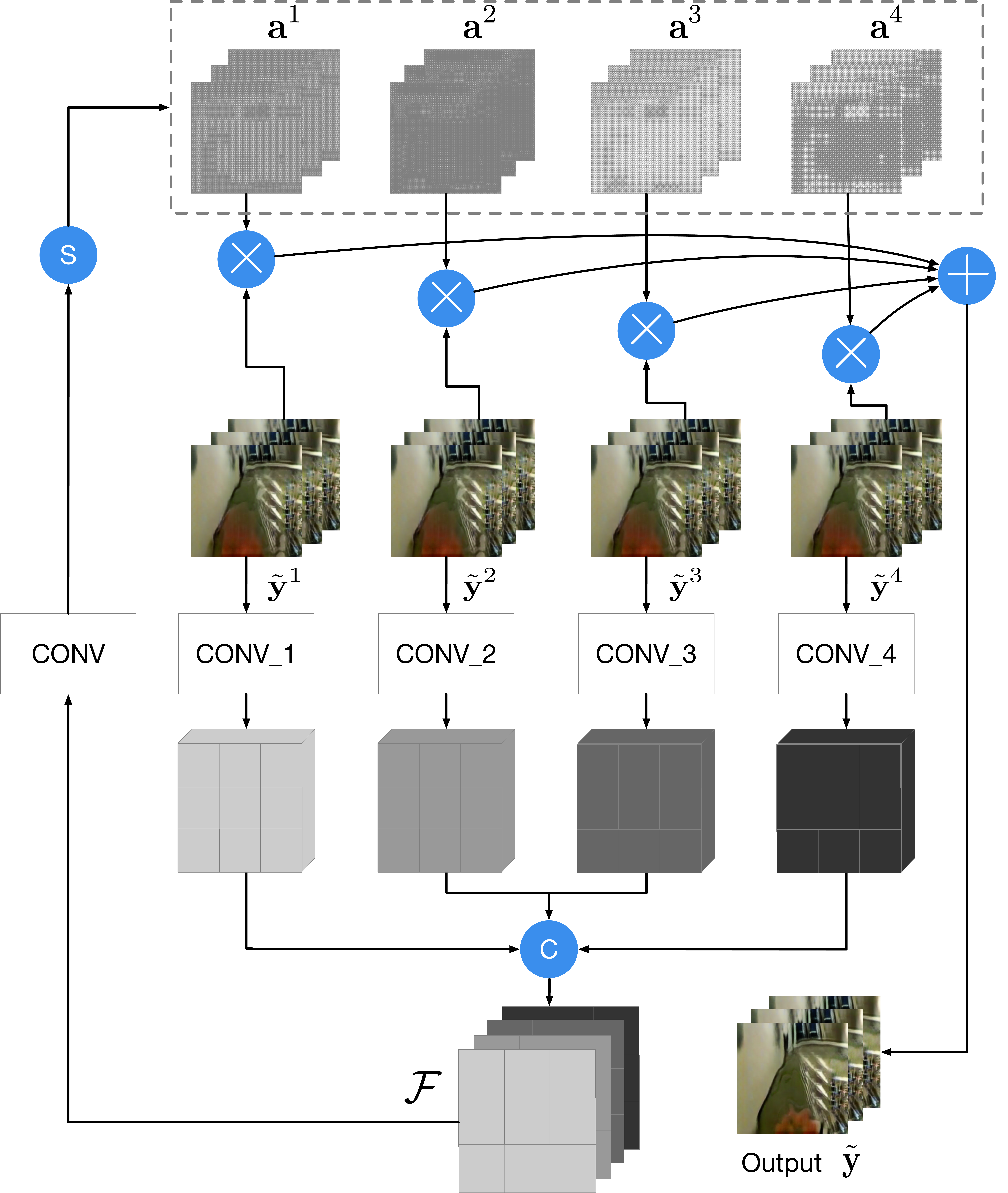}} 
\caption{Illustration of the proposed multi-generation attention fusion module. The symbol $\oplus$, $\otimes$, \textcircled{c} and \textcircled{s} denote element-wise addition, element-wise multiplication, channel-wise concatenation and channel-wise softmax operation.} 
	\vspace{-0.2cm}
\label{fig:attention}
\end{figure}

After the Bi-directional temporal and spatial modules, we obtain four generated video sequences $[\rm \tilde{\textbf{\textbf{y}}}^{1}, \rm \tilde{\textbf{\textbf{y}}}^{2}, \rm \tilde{\textbf{\textbf{y}}}^{3}, \rm \rm \tilde{\textbf{\textbf{y}}}^{4}]$. 
To combine relevant information from the four generated video sequences, we propose a novel multi-generation attention fusion module to obtain a refined video sequence. 
The details of the proposed multi-generation attention fusion module is shown in Figure~\ref{fig:attention}. 
First, four different features are extracted by four convolutional layers simultaneously from the four generated video sequences. 
Then the obtained four feature maps are concatenated to a new feature as:
\begin{equation}
\begin{aligned}
\mathcal{F} = \rm Concat(Conv(\rm \tilde{\textbf{\textbf{y}}}^{1}), Conv(\rm \tilde{\textbf{\textbf{y}}}^{2}), Conv(\rm \tilde{\textbf{\textbf{y}}}^{3}), Conv(\rm \tilde{\textbf{\textbf{y}}}^{4})),
\end{aligned}
\label{eqn:concat}
\end{equation}
where $\rm Concat(\cdot)$ and $\rm Conv(\cdot)$ denote channel-wise concatenation  operation and convolutional operation. 
Next, the concatenated feature $\mathcal{F}$ is fed into a de-convolutional layer to obtain the new size feature $\mathcal{F}'$ for attention fusion purpose. 
Then the attention maps $[\rm {\textbf{\textbf{a}}}^{1}, \rm {\textbf{\textbf{a}}}^{2}, \rm {\textbf{\textbf{a}}}^{3}, \rm {\textbf{\textbf{a}}}^{4}]$ are learned from a convolutional layer and softmax activation layer, where the softmax activation layer guarantees normalization of attention maps in channel-wise. Finally, the final video sequence $\rm \tilde{\textbf{\textbf{y}}}$ can be obtained as follows:
\begin{equation}
\begin{aligned}
\rm \tilde{\textbf{\textbf{y}}} = (\rm \tilde{\textbf{\textbf{y}}}^{1}\otimes \rm {\textbf{\textbf{a}}}^{1}) \oplus (\rm \tilde{\textbf{\textbf{y}}}^{2}\otimes \rm {\textbf{\textbf{a}}}^{2}) \oplus (\rm \tilde{\textbf{\textbf{y}}}^{3}\otimes \rm {\textbf{\textbf{a}}}^{3}) \oplus (\rm \tilde{\textbf{\textbf{y}}^4}\otimes \rm {\textbf{\textbf{a}}}^{4}),
\end{aligned}
\label{eqn:output}
\end{equation}
where $\rm \tilde{\textbf{\textbf{y}}}$ represents the final synthesized frame sequence, and the symbol $\otimes$ denotes the element-wise multiplication, and $\oplus$ is the element-wise addition.

Instead of multiplying feature maps and real images, we multiply feature maps $(\textbf{a}^1,\textbf{a}^2,\textbf{a}^3,\textbf{a}^4)$ and the intermediate results $(\tilde{\textbf{y}}^1, \tilde{\textbf{y}}^2, \tilde{\textbf{y}}^3, \tilde{\textbf{y}}^4)$ to obtain the final result $\tilde{\textbf{y}}$. We observe that a single-sequence generation space may not be suitable enough for learning a good mapping for this complex synthesis problem. Thus, we explore a larger sequence generation space to have a richer synthesis via constructing multiple intermediate sequence results. We first produce several intermediate sequences which can be regarded as the candidate sequence pool, then learn a set of attention maps. These attention maps are used to spatially select from the intermediate generations and are combined to synthesize the final output.

\begin{figure*}[htbp]
\centering
{\includegraphics[width=1\linewidth]{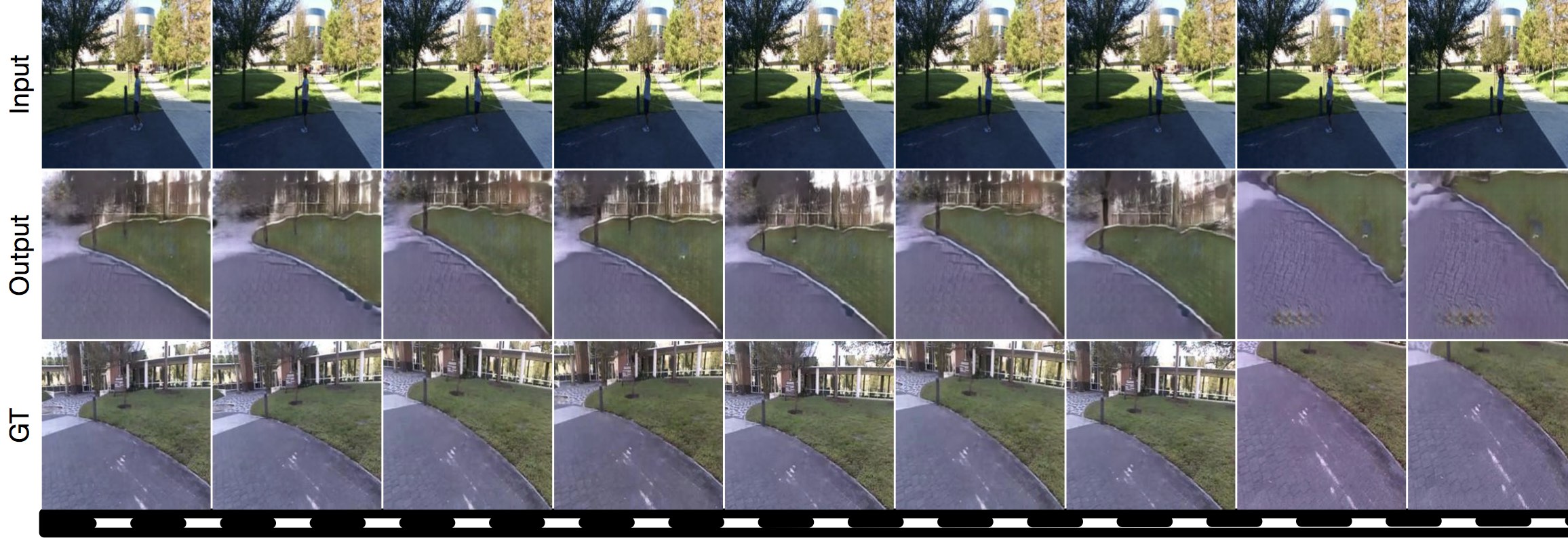}} 
\caption{Video frames generated from exocentric view to egocentric view on \textbf{Side2Ego} dataset.}
\label{fig:side2egoseq}
\end{figure*}

\begin{figure*}[htbp]
\centering
{\includegraphics[width=1\linewidth]{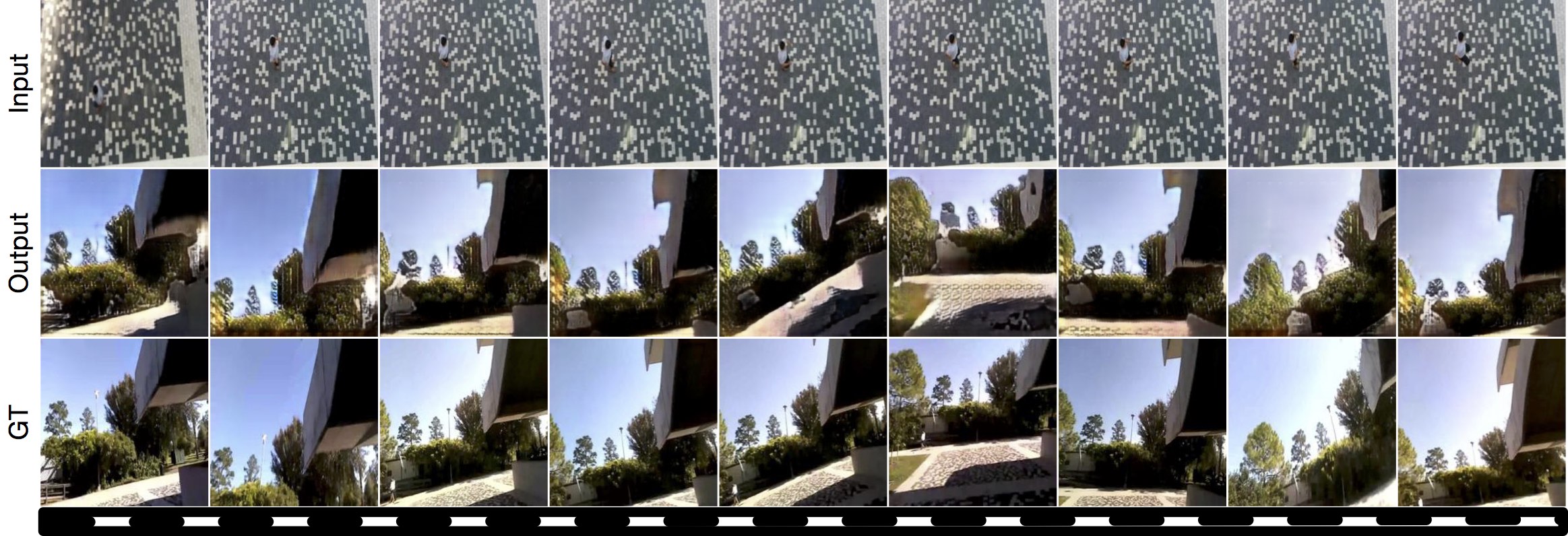}} 
\caption{Video frames generated from exocentric view to egocentric view on \textbf{Top2Ego} dataset.}
\label{fig:top2egoseq}
\end{figure*}

\subsection{Temporal and Spatial Dual-discriminator}
Traditional image-to-image generation methods use vanilla discriminator \cite{isola2017image}. 
In this paper, we propose a novel Temporal and Spatial Dual-discriminator, which contains two discriminators, i.e., temporal discriminator ($D_T$) and spatial discriminator ($D_S$). 
$D_T$ takes the real input $\rm x_1$/$\rm x_t$ and  the generated image sequence $\rm \tilde{\textbf{\textbf{y}}}$ as input. $D_T$ learns to tell whether a sequence of fake output $\rm \tilde{\textbf{\textbf{y}}}$ and input $\rm x_1$/$\rm x_{t}$ are related or not. Meanwhile, $D_S$ takes the real input image $\rm x $ and fake $\rm \tilde{y}$ as input. $D_S$ learns to tell whether the two frame sequences from different domains are associated with each other or not. $D_T$ and $D_S$ take temporal and spatial information into consideration, respectively.

Assuming we target to learn a mapping $G\colon \rm \textbf{x} \to \rm \tilde{\textbf{y}}$ from input exocentric view $\rm \textbf{x} {\equiv} {\{x_1,...,x_t\}}$ to output egocentric view $\rm \tilde{\textbf{y}} {\equiv} {\{\rm \tilde{y}_1,...,\rm \tilde{y}_t\}}$. The generator $G$ is trained to produce fake outputs $\rm \tilde{\textbf{y}}$ to fool the discriminator $D_T$ and $D_S$.
The adversarial loss can be expressed as:
\begin{equation}
\begin{aligned}
\mathcal{L}_{cGAN_S}(\rm \textbf{x},\rm \tilde{\textbf{y}}) =  
& \sum _{t=1}^t (\mathbb{E}_{\rm x_t, \rm y_t }\left[ \log D_S(\rm x_t, \rm y_t) \right] \\
& + \mathbb{E}_{\rm x_t, \rm \tilde y_t} \left[\log (1 - D_S(\rm x_t, \rm \tilde y_t)) \right]).
\end{aligned}
\label{eqn:advS}
\end{equation}

\begin{equation}
\begin{aligned}
\mathcal{L}_{cGAN_T}(\rm \textbf{x},\rm \tilde{\textbf{y}}) &= \mathbb{E}_{\rm {x}, \rm {\textbf{y}}} \left[ \log D_T(\rm {x_t}, \rm \textbf{y}) \right] \\
& + \mathbb{E}_{\rm {x}, \rm \tilde{\textbf{y}}} \left[\log (1 - D_T(\rm {x_t}, \rm \tilde{\textbf{y}})) \right] \\
& +  \mathbb{E}_{\rm {x}, \rm {\textbf{y}}} \left[ \log D_T(\rm {x_1}, \rm \textbf{y}) \right]\\
& + \mathbb{E}_{\rm {x}, \rm \tilde{\textbf{y}}} \left[\log (1 - D_T(\rm {x_1}, \rm \tilde{\textbf{y}})) \right].
\end{aligned}
\label{eqn:advT}
\end{equation}
$\rm x_1$/$\rm x_{t}$ in Equation~\eqref{eqn:advT} is the starting and ending frame in the temporal synthesis.
The total adversarial loss is formulated as follows:
\begin{equation}
\begin{aligned}
\mathcal{L}_{cGAN}(\rm \textbf{x},\rm \tilde{\textbf{y}}) =  \mathcal{L}_{cGAN_S} + \lambda_g\mathcal{L}_{cGAN_T}.
\end{aligned}
\label{eqn:gan}
\end{equation}

\subsection{Optimization Objective}
The training objective can be decomposed into four main components which are adversarial loss, temporal loss, spatial loss and reconstruction loss.

\noindent \textbf{Reconstruction Loss.} The task of the generator is to reconstruct an video sequences $\rm \tilde{\textbf{y}} {\equiv} {\{\rm \tilde{y}_1,...,\rm \tilde{y}_t\}}$ as close as the target sequences $\rm {\textbf{y}} {\equiv} {\{\rm {y}_1,...,\rm {y}_t\}}$. We use $\mathcal{L}1$ distance denoted as $\|\centerdot\|_1$ in the reconstruction loss:
\begin{equation}
\begin{aligned}
\mathcal{L}_{re}(G) =  
&\sum_{t=1}^t \lambda_r\mathbb{E}_{\rm \textbf{x},\rm \tilde{\textbf{y}}} \left[ \|\rm y_{t}-\rm \tilde{y}_t)\|_1 \right].
\end{aligned}
\label{eqn:rec}
\end{equation}

\noindent \textbf{Overall Loss.} The final optimization loss is a weighted sum of the above losses. Generator $G$ and discriminator $D$ are trained in an end-to-end fashion to optimize the following objective function:
\begin{equation}
\begin{aligned}
\mathcal{L} =  \mathcal{L}_{cGAN} + \mathcal{L}_{re} + \mathcal{L}_{T} + \mathcal{L}_{S},
\end{aligned}
\label{eqn:overall}
\end{equation}
where $\lambda_i$'s in Equations~\eqref{eqn:timedl}, \eqref{eqn:timeul}, \eqref{eqn:sdloss}, \eqref{eqn:suploss}, \eqref{eqn:gan}, \eqref{eqn:rec} are the regularization parameters.

\noindent \textbf{Network Architecture.}
We employ U-Net~\cite{Unet} as the architecture of our generator $G$. We impose the skip connection strategy from down-sampling path to up-sampling path to avoid the vanishing gradient problem. 
We adopt PatchGAN~\cite{isola2017image} for the discriminators $D_S$ and $D_T$. 
The feature maps for attention fusion are extracted by the up-sampling layers of the U-Net during the training of generator $G$. 
We adopt RefineNet~\cite{Lin:2017:RefineNet} to generate semantic maps on the Side2Ego and Top2Ego datasets as in~\cite{Regmi_2018_CVPR,tang2019multichannel}. 

\section{Experiments}

\begin{figure*}[htbp]
	\centering
	{\includegraphics[width=1\linewidth]{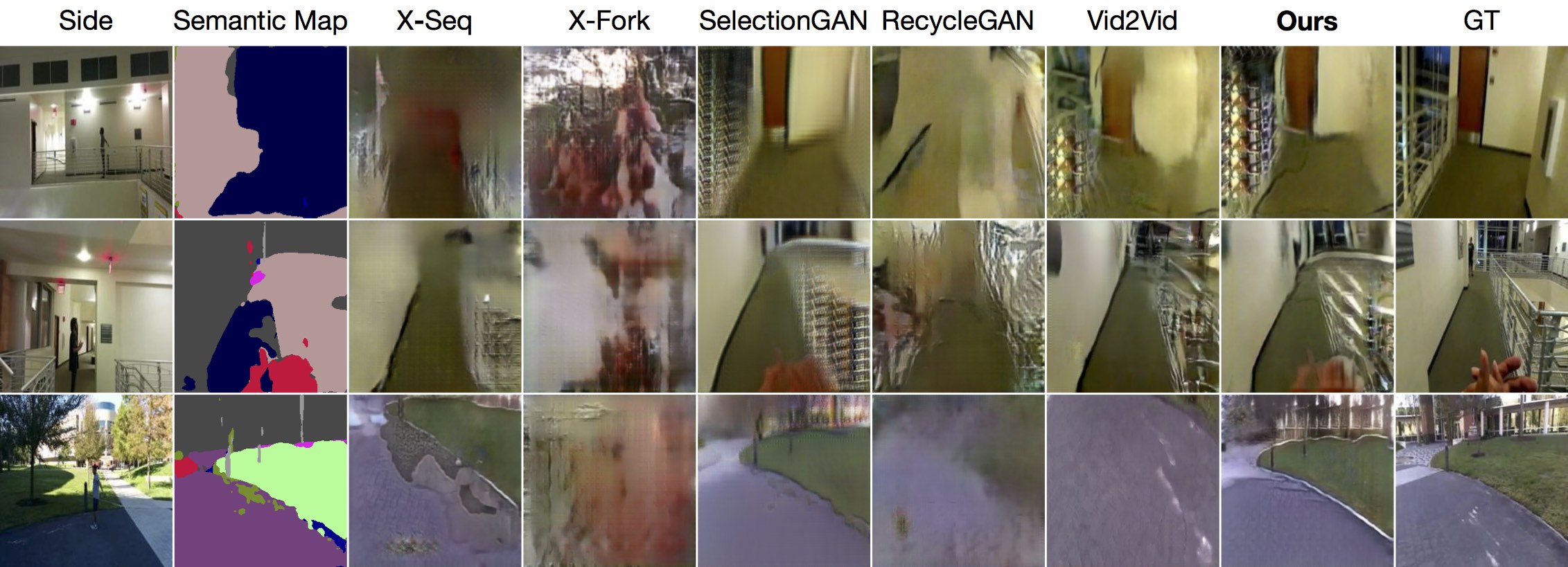}} 
	\caption{Video frames generated from exocentric view to egocentric view on \textbf{Side2Ego} dataset using different methods.}
	\label{fig:compareside}
\end{figure*}

\begin{table*}[htbp]
	\centering
	\caption{Quantitative evaluation of different image and video generation methods on the \textbf{Side2Ego} dataset. For these metrics except KL score and FID, higher is better.}
	\begin{tabular}{rccccccccc} \toprule 
		\multirow{2}{*}{Method}  &  \multirow{2}{*}{SSIM $\uparrow$} & \multirow{2}{*}{PSNR $\uparrow$} & \multirow{2}{*}{SD $\uparrow$} & \multirow{2}{*}{KL $\downarrow$} & \multirow{2}{*}{FID $\downarrow$} & \multicolumn{2}{c}{Top-1 $\uparrow$} & \multicolumn{2}{c}{Top-5 $\uparrow$} 
		\\&&&&&& \multicolumn{2}{c}{Accuracy (\%)} & \multicolumn{2}{c}{Accuracy (\%)}  \\ \midrule
		X-Fork \cite{Regmi_2018_CVPR}         &0.4499 & 17.0743 & 20.4443 & 51.20 $\pm$ 1.94 & 216.5575 & 4.49 & 9.76 & 11.63& 19.44 \\
		X-Seq \cite{Regmi_2018_CVPR}  & 0.4763 & 17.1462 & 20.7468 & 45.10 $\pm$ 1.95 & 184.2808&  6.51 & 12.70 & 11.97 & 19.36\\
		SelectionGAN \cite{tang2019multichannel}  & 0.5128 & 18.3021 & 20.9426 & \textbf{7.26 $\pm$ 1.27} &\textbf{139.1429} & 20.84 & 37.49 & 42.51 & 65.22\\
		RecycleGAN \cite{Recycle-GAN}  &0.3446& 15.9242  & 18.9429& 42.40 $\pm$1.61 &186.5897 & 2.32 &2.40 & 9.13 & 10.98  \\
		Vid2Vid \cite{wang2018vid2vid}  & 0.3955 & 15.9012 & 19.7169 & 59.41 $\pm$ 1.93 & 196.9749 & 7.52 & 14.70 & 13.97 & 24.36 \\ \hline
		STA-GAN (Ours) & \textbf{0.5607} & \textbf{20.7027} & \textbf{20.9491} & 9.44 $\pm$ 1.48 & 169.3514 & \textbf{26.83} &\textbf{39.83}&\textbf{42.72} &\textbf{69.30}\\
		\bottomrule  
	\end{tabular}
	\label{tab:side}
\end{table*}

\begin{table*}[htbp]
	\centering
	\caption{Quantitative evaluation of different image and video generation methods on the \textbf{Top2Ego} dataset. For these metrics except KL score and FID, higher is better.}
	\begin{tabular}{rccccccccc} \toprule
		\multirow{2}{*}{Method}  &  \multirow{2}{*}{SSIM $\uparrow$} & \multirow{2}{*}{PSNR $\uparrow$} & \multirow{2}{*}{SD $\uparrow$} & \multirow{2}{*}{KL $\downarrow$} & \multirow{2}{*}{FID $\downarrow$} &  \multicolumn{2}{c}{Top-1 $\uparrow$} & \multicolumn{2}{c}{Top-5 $\uparrow$} 
		\\&&&&&& \multicolumn{2}{c}{Accuracy (\%)} & \multicolumn{2}{c}{Accuracy (\%)}  \\ \midrule
		X-Fork \cite {Regmi_2018_CVPR} &0.2952 & 15.8849 & 18.7349 & 63.96$\pm$1.74 & 208.6632 & 0.8 & 1.22 & 3.16 &4.08 \\
		X-Seq \cite {Regmi_2018_CVPR}  &0.3522 & 16.9439 & 19.2733 & 54.91 $\pm$ 1.81 & 198.3948 & 1.07 & 1.77 & 4.29 & 6.94\\ 
		SelectionGAN \cite{tang2019multichannel}  & 0.5047 & 22.0244 & 19.1976 & 10.07 $\pm$ 1.29 &199.1441 & 8.85 & 16.55 & 24.32 & 33.90\\
		RecycleGAN \cite{Recycle-GAN}  & 0.3264 & 16.4767 & 19.4543 & 31.55 $\pm$1.32 & 235.8220 & 0.43 & 2.04 &2.47  & 3.70   \\
		Vid2Vid \cite{wang2018vid2vid}  & 0.3895 & 17.1233 & 19.6043 & 31.48$\pm$1.77 & 208.9400 & 5.65 & 9.77 & 12.38 & 18.97  \\ \hline
		STA-GAN (Ours)  & \textbf{0.5383} & \textbf{22.5816} & \textbf{19.2895} & \textbf{10.02 $\pm$1.30 } & \textbf{175.7446} & \textbf{8.93} & \textbf{24.80} &\textbf{26.37} & \textbf{46.74}\\ \bottomrule  
	\end{tabular}
	\label{tab:top}
\end{table*}

\begin{figure*}[htbp] 
	\centering
	{\includegraphics[width=1\linewidth]{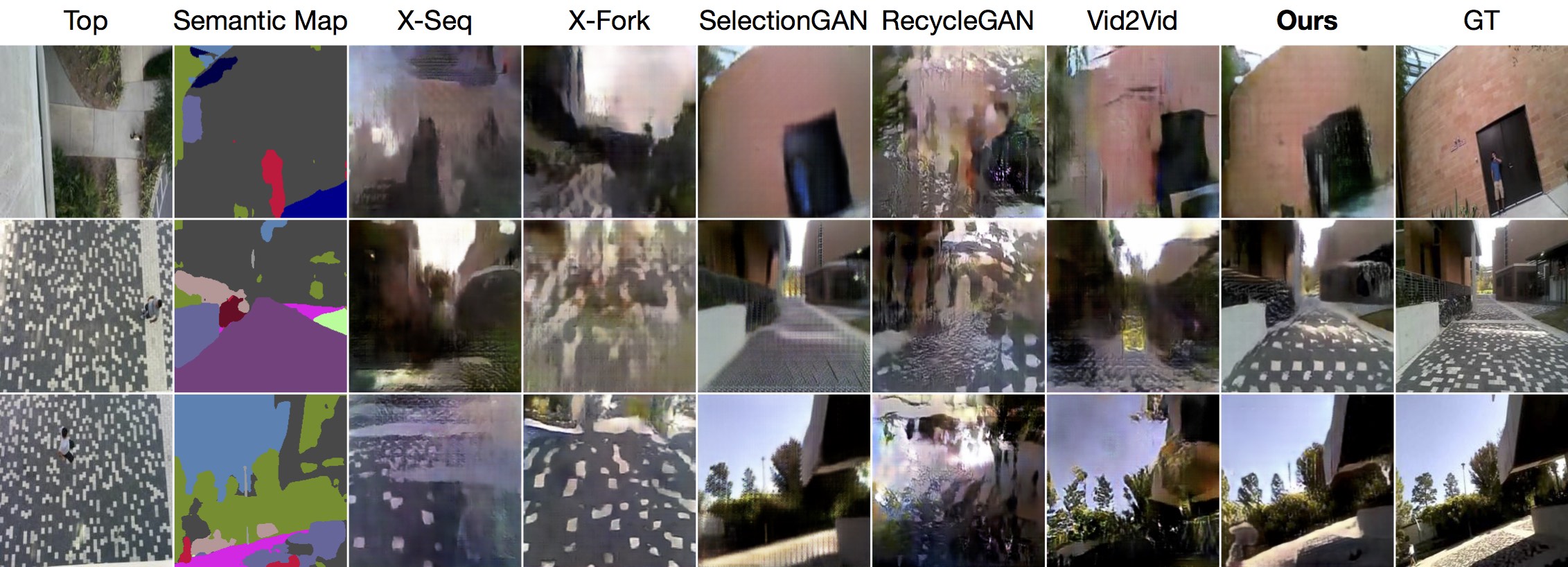}}
	\caption{Video frames generated from exocentric view to egocentric view on \textbf{Top2Ego} dataset using different methods.}
	\label{fig:comparetop}
\end{figure*}

\noindent \textbf{Datasets.}
To explore the effectiveness of the proposed STA-GAN, we conduct extensive experiments on the Side2Ego and Top2Ego datasets~\cite{third2019}. 
These datasets simultaneously recorded egocentric and exocentric videos. 
Each video pair contains one egocentric and one exocentric video (side-view or top-view). 
The pair of videos are temporally aligned. 
These datasets are challenging due to two reasons. First, it contains dramatically different indoor and outdoor scenes. 
Second, the datasets are collected simultaneously by an exocentric camera (side or top view) and an egocentric wearable camera. The datasets include a huge amount of blurred image frames for egocentric view. 
For Side2Ego dataset, there are 124 videos containing 26,764 pairs of frames for training and 13,788 pairs for testing. 
For Top2Ego dataset, there are 135 videos containing 28,408 pairs of frames for training and 14,064 pairs for testing. All image frames are in high-resolution with $1280 {\times} 720$ pixels. 

\noindent \textbf{Parameter Settings.}
All images are scaled to $256 {\times} 256$. 
We enable image flipping and random crops for data augmentation. 
We train 200 epochs with the batch size of 8. 
In our experiments, we set $\lambda_u {=} 1$, $\lambda_d {=} 0.1$, $\lambda_n {=} 1$, $\lambda_p {=} 0.1$, $\lambda_g {=} 10$, $\lambda_r {=} 10$ in Equations~\eqref{eqn:timedl}, \eqref{eqn:timeul}, \eqref{eqn:sdloss}, \eqref{eqn:suploss}, \eqref{eqn:gan}, \eqref{eqn:rec} respectively. The number of time truncat $i$ in Eq.~\eqref{eqn:sdn} and \eqref{eqn:sup} is set to 3. 
The proposed STA-GAN is implemented by PyTorch. We perform our experiments on NVIDIA Geforce GTX 1080 Ti GPU with 11 GB memory to accelerate training process.


\noindent \textbf{Evaluation Metrics.} 
We follow~\cite{wang2018vid2vid,Regmi_2018_CVPR,tang2019multichannel} and apply metrics such as top-k prediction accuracy, KL score and Fr\'echet Inception Distance (FID) for evaluating the proposed method. 
These metrics evaluate the generated images in a high-level feature space.
We also employ pixel-level similarity metrics in the experiments, i.e., Structural-Similarity (SSIM) \cite{Wang03}, Peak Signal-to-Noise Ratio (PSNR) and Sharpness Difference (SD) \cite{sharpness}. 

\subsection{State-of-the-Art Comparisons}
\noindent \textbf{Quantitative Comparisons.}
We compare our STA-GAN with both cross-view image generation methods, i.e., X-Fork~\cite{Regmi_2018_CVPR}, X-Seq~\cite{Regmi_2018_CVPR}, SelectionGAN~\cite{tang2019multichannel}, and video generation methods, i.e., RecycleGAN~\cite{Recycle-GAN} and Vid2Vid~\cite{wang2018vid2vid}. 
The quantitative results compared with state-of-the-art methods on the Side2Ego and Top2Ego datasets are presented in Table~\ref{tab:side} and  Table~\ref{tab:top}. 
We observe that the proposed STA-GAN achieves better results than state-of-the-art methods in most cases. Accuracies are computed in two ways, the first column considers all images, whereas the second column computes accuracy of real images whose top-1 prediction is greater than 0.5. Particularly, we increase the metric PSNR by 2.4, Top-1 accuracy by 6\%/2\%, Top-5 accuracy by 4\%/5\%, compared to the second best baseline on Side2Ego dataset. Meanwhile, we increase the metric PSNR by 0.5, Top-1 accuracy by 0.1\%/8\%, Top-5 accuracy by 2\%/13\%, compared with the second best baseline on Top2Ego dataset. These comparisons show the effectiveness of STA-GAN for cross-view exocentric to egocentric video synthesis.

\noindent \textbf{Qualitative Comparisons.}
Video sequence synthesis results are shown in Figure~\ref{fig:side2egoseq} and Figure~\ref{fig:top2egoseq}. We can observe that the synthesized egocentric video sequence change smoothly from left to right, and are visually close to the ground truth. The qualitative results compared with state-of-the-art methods are shown in Figure~\ref{fig:compareside} and Figure~\ref{fig:comparetop}. We observe that our method generates more clear and reasonable video frames than other methods. 
It is obvious that objects are in the correct positions for generated egocentric video frames using our method. 
Moreover, the structure and layout of the generated video frames are closer to the ground truth. 
Results show that egocentric video frames generated by STA-GAN are visually better compared with other baselines, which further confirm that the proposed STA-GAN network has the ability to transfer the video sequences from exocentric to egocentric perspective.

\subsection{Ablation Study}
To evaluate the performance of proposed STA-GAN, we conduct experiments with different settings on the Top2Ego dataset. 
As shown in Table~\ref{tab:ablation}, the proposed STA-GAN considers six different settings.  
Baseline A utilizes only spatial information during generation process while baseline B utilizes only temporal information. 
We observe that baseline B performs better than baseline A which demonstrates that temporal information is more important for video generation task. 
Baseline C by considering both temporal and spatial information improves the SSIM, PSNR and SD metrics to 0.3098/17.0236/18.6043 respectively, meaning that both spatial and temporal information should be explored for video generation task. 
Baseline D by incorporating both bi-directional downstream and upstream generation further improves the performance to 0.4287/20.2891/19.2389, which demonstrates that generation are reversible in video generation task.
Baseline E outperforms D showing the importance of using the proposed dual-discriminator, i.e., temporal discriminator and spatial discriminator.
Baseline F by adopting the attention fusion strategy further increases the SSIM, PSNR and SD scores, which demonstrates the effectiveness of attention fusion. 

\begin{table}[!tbp]
	\centering
	\caption{Ablation study of STA-GAN on the \textbf{Top2Ego} dataset.}
		\begin{tabular}{clccc} 
			\toprule
			No. & Setting of STA-GAN & SSIM $\uparrow$ & PSNR $\uparrow$ & SD $\uparrow$ \\
			\midrule
			A & Spatial Generation & 0.2568 &15.8561 & 18.1414 \\ 
			B & Temporal Generation & 0.2627 & 15.3411 & 18.1914 \\
			C & Spatial + Temporal  &  0.3098 & 17.0236 & 18.6043 \\
			D & C + Bi-Direction & 0.4287 & 20.2891 & 19.2389 \\
			E & D + Dual-Discriminator & 0.4956 & \textbf{23.4734} & 19.2526\\
			F & E + Attention Module & \textbf{0.5383} & 22.5816 & \textbf{19.2895} \\ 
			\bottomrule  
	\end{tabular}
	\label{tab:ablation}
\end{table}
\section{Conclusion}
In this paper, we propose a novel STA-GAN framework to address a novel cross-view  exocentric to egocentric video synthesize problem by exploiting the temporal and spatial information in videos. Based on the property of videos, we propose a bi-directional strategy which generates video sequences in both downstream and upstream directions. Meanwhile, a novel  temporal and spatial dual-discriminator is proposed for better network training.
Moreover, we propose a novel attention fusion method which targets to refine the generation results.
Extensive experimental results on the Top2Ego and Side2Ego datasets demonstrate that our method outperforms state-of-the-art approaches for the challenging cross-view exocentric to egocentric video synthesis.

\bibliographystyle{ACM-Reference-Format}
\bibliography{sample-base}


\begin{thebibliography}{57}


\ifx \showCODEN    \undefined \def \showCODEN     #1{\unskip}     \fi
\ifx \showDOI      \undefined \def \showDOI       #1{#1}\fi
\ifx \showISBNx    \undefined \def \showISBNx     #1{\unskip}     \fi
\ifx \showISBNxiii \undefined \def \showISBNxiii  #1{\unskip}     \fi
\ifx \showISSN     \undefined \def \showISSN      #1{\unskip}     \fi
\ifx \showLCCN     \undefined \def \showLCCN      #1{\unskip}     \fi
\ifx \shownote     \undefined \def \shownote      #1{#1}          \fi
\ifx \showarticletitle \undefined \def \showarticletitle #1{#1}   \fi
\ifx \showURL      \undefined \def \showURL       {\relax}        \fi
\providecommand\bibfield[2]{#2}
\providecommand\bibinfo[2]{#2}
\providecommand\natexlab[1]{#1}
\providecommand\showeprint[2][]{arXiv:#2}

\bibitem[\protect\citeauthoryear{??}{Gop}{[n.d.]}]%
        {Gopro}
 \bibinfo{year}{[n.d.]}\natexlab{}.
\newblock \bibinfo{howpublished}{\url{https://gopro.com/en/us/}}.
\newblock


\bibitem[\protect\citeauthoryear{Aghazadeh, Sullivan, and Carlsson}{Aghazadeh
  et~al\mbox{.}}{2011}]%
        {omid}
\bibfield{author}{\bibinfo{person}{Omid Aghazadeh}, \bibinfo{person}{Josephine
  Sullivan}, {and} \bibinfo{person}{Stefan Carlsson}.}
  \bibinfo{year}{2011}\natexlab{}.
\newblock \showarticletitle{Novelty detection from an ego-centric perspective}.
  In \bibinfo{booktitle}{\emph{CVPR}}.
\newblock


\bibitem[\protect\citeauthoryear{Ardeshir and Borji}{Ardeshir and
  Borji}{2016}]%
        {ego2top}
\bibfield{author}{\bibinfo{person}{Shervin Ardeshir} {and} \bibinfo{person}{Ali
  Borji}.} \bibinfo{year}{2016}\natexlab{}.
\newblock \showarticletitle{Ego2top: Matching viewers in egocentric and
  top-view videos}. In \bibinfo{booktitle}{\emph{ECCV}}.
\newblock


\bibitem[\protect\citeauthoryear{Bansal, Ma, Ramanan, and Sheikh}{Bansal
  et~al\mbox{.}}{2018}]%
        {Recycle-GAN}
\bibfield{author}{\bibinfo{person}{Aayush Bansal}, \bibinfo{person}{Shugao Ma},
  \bibinfo{person}{Deva Ramanan}, {and} \bibinfo{person}{Yaser Sheikh}.}
  \bibinfo{year}{2018}\natexlab{}.
\newblock \showarticletitle{Recycle-gan: Unsupervised video retargeting}. In
  \bibinfo{booktitle}{\emph{ECCV}}.
\newblock


\bibitem[\protect\citeauthoryear{Brock, Donahue, and Simonyan}{Brock
  et~al\mbox{.}}{2019}]%
        {brock2018large}
\bibfield{author}{\bibinfo{person}{Andrew Brock}, \bibinfo{person}{Jeff
  Donahue}, {and} \bibinfo{person}{Karen Simonyan}.}
  \bibinfo{year}{2019}\natexlab{}.
\newblock \showarticletitle{Large scale GAN training for high fidelity natural
  image synthesis}. In \bibinfo{booktitle}{\emph{ICLR}}.
\newblock


\bibitem[\protect\citeauthoryear{Chan, Ginosar, Zhou, and Efros}{Chan
  et~al\mbox{.}}{2019}]%
        {chan2019dance}
\bibfield{author}{\bibinfo{person}{Caroline Chan}, \bibinfo{person}{Shiry
  Ginosar}, \bibinfo{person}{Tinghui Zhou}, {and} \bibinfo{person}{Alexei~A
  Efros}.} \bibinfo{year}{2019}\natexlab{}.
\newblock \showarticletitle{Everybody dance now}. In
  \bibinfo{booktitle}{\emph{ICCV}}.
\newblock


\bibitem[\protect\citeauthoryear{Chen, Wang, Yang, Zhang, Xiong, Loy, and
  Lin}{Chen et~al\mbox{.}}{2018}]%
        {STLattice2018CVPR}
\bibfield{author}{\bibinfo{person}{Kai Chen}, \bibinfo{person}{Jiaqi Wang},
  \bibinfo{person}{Shuo Yang}, \bibinfo{person}{Xingcheng Zhang},
  \bibinfo{person}{Yuanjun Xiong}, \bibinfo{person}{Chen~Change Loy}, {and}
  \bibinfo{person}{Dahua Lin}.} \bibinfo{year}{2018}\natexlab{}.
\newblock \showarticletitle{Optimizing video object detection via a scale-time
  lattice}. In \bibinfo{booktitle}{\emph{CVPR}}.
\newblock


\bibitem[\protect\citeauthoryear{Choi, Uh, Yoo, and Ha}{Choi
  et~al\mbox{.}}{2020}]%
        {choi2019stargan}
\bibfield{author}{\bibinfo{person}{Yunjey Choi}, \bibinfo{person}{Youngjung
  Uh}, \bibinfo{person}{Jaejun Yoo}, {and} \bibinfo{person}{Jung-Woo Ha}.}
  \bibinfo{year}{2020}\natexlab{}.
\newblock \showarticletitle{Stargan v2: Diverse image synthesis for multiple
  domains}. In \bibinfo{booktitle}{\emph{CVPR}}.
\newblock


\bibitem[\protect\citeauthoryear{Denton and Fergus}{Denton and Fergus}{2018}]%
        {stovid}
\bibfield{author}{\bibinfo{person}{Emily Denton} {and} \bibinfo{person}{Rob
  Fergus}.} \bibinfo{year}{2018}\natexlab{}.
\newblock \showarticletitle{Stochastic video generation with a learned prior}.
  In \bibinfo{booktitle}{\emph{ICML}}.
\newblock


\bibitem[\protect\citeauthoryear{Duan, Wang, Tang, Latapie, and Yan}{Duan
  et~al\mbox{.}}{2021}]%
        {duan2021cascade}
\bibfield{author}{\bibinfo{person}{Bin Duan}, \bibinfo{person}{Wei Wang},
  \bibinfo{person}{Hao Tang}, \bibinfo{person}{Hugo Latapie}, {and}
  \bibinfo{person}{Yan Yan}.} \bibinfo{year}{2021}\natexlab{}.
\newblock \showarticletitle{Cascade attention guided residue learning gan for
  cross-modal translation}. In \bibinfo{booktitle}{\emph{ICPR}}.
\newblock


\bibitem[\protect\citeauthoryear{Elfeki, Regmi, Ardeshir, and Borji}{Elfeki
  et~al\mbox{.}}{2019}]%
        {third2019}
\bibfield{author}{\bibinfo{person}{Mohamed Elfeki}, \bibinfo{person}{Krishna
  Regmi}, \bibinfo{person}{Shervin Ardeshir}, {and} \bibinfo{person}{Ali
  Borji}.} \bibinfo{year}{2019}\natexlab{}.
\newblock \showarticletitle{From third person to first person: Dataset and
  baselines for synthesis and retrieval}. In \bibinfo{booktitle}{\emph{CVPR}}.
\newblock


\bibitem[\protect\citeauthoryear{Fan, Lee, Xu, Kumar~Singh, Jae~Lee, Crandall,
  and Ryoo}{Fan et~al\mbox{.}}{2017}]%
        {Xu17}
\bibfield{author}{\bibinfo{person}{Chenyou Fan}, \bibinfo{person}{Jangwon Lee},
  \bibinfo{person}{Mingze Xu}, \bibinfo{person}{Krishna Kumar~Singh},
  \bibinfo{person}{Yong Jae~Lee}, \bibinfo{person}{David~J Crandall}, {and}
  \bibinfo{person}{Michael~S Ryoo}.} \bibinfo{year}{2017}\natexlab{}.
\newblock \showarticletitle{Identifying first-person camera wearers in
  third-person videos}. In \bibinfo{booktitle}{\emph{CVPR}}.
\newblock


\bibitem[\protect\citeauthoryear{Fathi, Farhadi, and Rehg}{Fathi
  et~al\mbox{.}}{2011}]%
        {Fathi11}
\bibfield{author}{\bibinfo{person}{Alireza Fathi}, \bibinfo{person}{Ali
  Farhadi}, {and} \bibinfo{person}{James~M Rehg}.}
  \bibinfo{year}{2011}\natexlab{}.
\newblock \showarticletitle{Understanding egocentric activities}. In
  \bibinfo{booktitle}{\emph{ICCV}}.
\newblock


\bibitem[\protect\citeauthoryear{Fathi, Hodgins, and Rehg}{Fathi
  et~al\mbox{.}}{2012a}]%
        {FathiCVPR2012}
\bibfield{author}{\bibinfo{person}{Alircza Fathi}, \bibinfo{person}{Jessica~K
  Hodgins}, {and} \bibinfo{person}{James~M Rehg}.}
  \bibinfo{year}{2012}\natexlab{a}.
\newblock \showarticletitle{Social interactions: A first-person perspective}.
  In \bibinfo{booktitle}{\emph{CVPR}}.
\newblock


\bibitem[\protect\citeauthoryear{Fathi, Li, and Rehg}{Fathi
  et~al\mbox{.}}{2012b}]%
        {FathiECCV2012}
\bibfield{author}{\bibinfo{person}{Alireza Fathi}, \bibinfo{person}{Yin Li},
  {and} \bibinfo{person}{James~M Rehg}.} \bibinfo{year}{2012}\natexlab{b}.
\newblock \showarticletitle{Learning to recognize daily actions using gaze}. In
  \bibinfo{booktitle}{\emph{ECCV}}.
\newblock


\bibitem[\protect\citeauthoryear{Goodfellow, Pouget-Abadie, Mirza, Xu,
  Warde-Farley, Ozair, Courville, and Bengio}{Goodfellow et~al\mbox{.}}{2014}]%
        {NIPS2014_5423}
\bibfield{author}{\bibinfo{person}{Ian Goodfellow}, \bibinfo{person}{Jean
  Pouget-Abadie}, \bibinfo{person}{Mehdi Mirza}, \bibinfo{person}{Bing Xu},
  \bibinfo{person}{David Warde-Farley}, \bibinfo{person}{Sherjil Ozair},
  \bibinfo{person}{Aaron Courville}, {and} \bibinfo{person}{Yoshua Bengio}.}
  \bibinfo{year}{2014}\natexlab{}.
\newblock \showarticletitle{Generative adversarial nets}.
\newblock  (\bibinfo{year}{2014}).
\newblock


\bibitem[\protect\citeauthoryear{Huh, Sun, and Zhang}{Huh
  et~al\mbox{.}}{2019}]%
        {huh2019feedback}
\bibfield{author}{\bibinfo{person}{Minyoung Huh}, \bibinfo{person}{Shao-Hua
  Sun}, {and} \bibinfo{person}{Ning Zhang}.} \bibinfo{year}{2019}\natexlab{}.
\newblock \showarticletitle{Feedback adversarial learning: Spatial feedback for
  improving generative adversarial networks}. In
  \bibinfo{booktitle}{\emph{CVPR}}.
\newblock


\bibitem[\protect\citeauthoryear{Isola, Zhu, Zhou, and Efros}{Isola
  et~al\mbox{.}}{2017}]%
        {isola2017image}
\bibfield{author}{\bibinfo{person}{Phillip Isola}, \bibinfo{person}{Jun-Yan
  Zhu}, \bibinfo{person}{Tinghui Zhou}, {and} \bibinfo{person}{Alexei~A
  Efros}.} \bibinfo{year}{2017}\natexlab{}.
\newblock \showarticletitle{Image-to-image translation with conditional
  adversarial networks}. In \bibinfo{booktitle}{\emph{CVPR}}.
\newblock


\bibitem[\protect\citeauthoryear{Kanade and Hebert}{Kanade and Hebert}{2012}]%
        {firstperson_kanade}
\bibfield{author}{\bibinfo{person}{Takeo Kanade} {and} \bibinfo{person}{Martial
  Hebert}.} \bibinfo{year}{2012}\natexlab{}.
\newblock \showarticletitle{First-person vision}.
\newblock \bibinfo{journal}{\emph{Proc. IEEE}} \bibinfo{volume}{100},
  \bibinfo{number}{8} (\bibinfo{year}{2012}), \bibinfo{pages}{2442--2453}.
\newblock


\bibitem[\protect\citeauthoryear{Karras, Laine, and Aila}{Karras
  et~al\mbox{.}}{2019}]%
        {karras2019style}
\bibfield{author}{\bibinfo{person}{Tero Karras}, \bibinfo{person}{Samuli
  Laine}, {and} \bibinfo{person}{Timo Aila}.} \bibinfo{year}{2019}\natexlab{}.
\newblock \showarticletitle{A style-based generator architecture for generative
  adversarial networks}. In \bibinfo{booktitle}{\emph{CVPR}}.
\newblock


\bibitem[\protect\citeauthoryear{Lin, Liu, Milan, Shen, and Reid}{Lin
  et~al\mbox{.}}{2019}]%
        {lin2019refinenet}
\bibfield{author}{\bibinfo{person}{Guosheng Lin}, \bibinfo{person}{Fayao Liu},
  \bibinfo{person}{Anton Milan}, \bibinfo{person}{Chunhua Shen}, {and}
  \bibinfo{person}{Ian Reid}.} \bibinfo{year}{2019}\natexlab{}.
\newblock \showarticletitle{Refinenet: Multi-path refinement networks for dense
  prediction}.
\newblock \bibinfo{journal}{\emph{IEEE TPAMI}} \bibinfo{volume}{42},
  \bibinfo{number}{5} (\bibinfo{year}{2019}), \bibinfo{pages}{1228--1242}.
\newblock


\bibitem[\protect\citeauthoryear{Lin, Milan, Shen, and Reid}{Lin
  et~al\mbox{.}}{2017}]%
        {Lin:2017:RefineNet}
\bibfield{author}{\bibinfo{person}{Guosheng Lin}, \bibinfo{person}{Anton
  Milan}, \bibinfo{person}{Chunhua Shen}, {and} \bibinfo{person}{Ian Reid}.}
  \bibinfo{year}{2017}\natexlab{}.
\newblock \showarticletitle{Refinenet: Multi-path refinement networks for
  high-resolution semantic segmentation}. In \bibinfo{booktitle}{\emph{CVPR}}.
\newblock


\bibitem[\protect\citeauthoryear{Liu, Tang, Latapie, and Yan}{Liu
  et~al\mbox{.}}{2020}]%
        {liu2020exocentric}
\bibfield{author}{\bibinfo{person}{Gaowen Liu}, \bibinfo{person}{Hao Tang},
  \bibinfo{person}{Hugo Latapie}, {and} \bibinfo{person}{Yan Yan}.}
  \bibinfo{year}{2020}\natexlab{}.
\newblock \showarticletitle{Exocentric to egocentric image generation via
  parallel generative adversarial network}. In
  \bibinfo{booktitle}{\emph{ICASSP}}.
\newblock


\bibitem[\protect\citeauthoryear{Liu, Huang, Mallya, Karras, Aila, Lehtinen,
  and Kautz}{Liu et~al\mbox{.}}{2019}]%
        {liu2019few}
\bibfield{author}{\bibinfo{person}{Ming-Yu Liu}, \bibinfo{person}{Xun Huang},
  \bibinfo{person}{Arun Mallya}, \bibinfo{person}{Tero Karras},
  \bibinfo{person}{Timo Aila}, \bibinfo{person}{Jaakko Lehtinen}, {and}
  \bibinfo{person}{Jan Kautz}.} \bibinfo{year}{2019}\natexlab{}.
\newblock \showarticletitle{Few-shot unsupervised image-to-image translation}.
  In \bibinfo{booktitle}{\emph{ICCV}}.
\newblock


\bibitem[\protect\citeauthoryear{Liu and Tuzel}{Liu and Tuzel}{2016}]%
        {couplegan}
\bibfield{author}{\bibinfo{person}{Ming-Yu Liu} {and} \bibinfo{person}{Oncel
  Tuzel}.} \bibinfo{year}{2016}\natexlab{}.
\newblock \showarticletitle{Coupled generative adversarial networks}.
\newblock \bibinfo{journal}{\emph{NeurIPS}} (\bibinfo{year}{2016}).
\newblock


\bibitem[\protect\citeauthoryear{Mathieu, Couprie, and LeCun}{Mathieu
  et~al\mbox{.}}{2016}]%
        {sharpness}
\bibfield{author}{\bibinfo{person}{Michael Mathieu}, \bibinfo{person}{Camille
  Couprie}, {and} \bibinfo{person}{Yann LeCun}.}
  \bibinfo{year}{2016}\natexlab{}.
\newblock \showarticletitle{Deep multi-scale video prediction beyond mean
  square error}. In \bibinfo{booktitle}{\emph{ICLR}}.
\newblock


\bibitem[\protect\citeauthoryear{Ogaki, Kitani, Sugano, and Sato}{Ogaki
  et~al\mbox{.}}{2012}]%
        {ogaki2012coupling}
\bibfield{author}{\bibinfo{person}{Keisuke Ogaki}, \bibinfo{person}{Kris~M
  Kitani}, \bibinfo{person}{Yusuke Sugano}, {and} \bibinfo{person}{Yoichi
  Sato}.} \bibinfo{year}{2012}\natexlab{}.
\newblock \showarticletitle{Coupling eye-motion and ego-motion features for
  first-person activity recognition}. In \bibinfo{booktitle}{\emph{CVPR
  Workshops}}.
\newblock


\bibitem[\protect\citeauthoryear{Pan, Wang, Jia, Shao, Sheng, Yan, and
  Wang}{Pan et~al\mbox{.}}{2019}]%
        {seg2vid_pan}
\bibfield{author}{\bibinfo{person}{Junting Pan}, \bibinfo{person}{Chengyu
  Wang}, \bibinfo{person}{Xu Jia}, \bibinfo{person}{Jing Shao},
  \bibinfo{person}{Lu Sheng}, \bibinfo{person}{Junjie Yan}, {and}
  \bibinfo{person}{Xiaogang Wang}.} \bibinfo{year}{2019}\natexlab{}.
\newblock \showarticletitle{Video generation from single semantic label map}.
  In \bibinfo{booktitle}{\emph{CVPR}}.
\newblock


\bibitem[\protect\citeauthoryear{Park, Yang, Yumer, Ceylan, and Berg}{Park
  et~al\mbox{.}}{2017}]%
        {tvsn_cvpr2017}
\bibfield{author}{\bibinfo{person}{Eunbyung Park}, \bibinfo{person}{Jimei
  Yang}, \bibinfo{person}{Ersin Yumer}, \bibinfo{person}{Duygu Ceylan}, {and}
  \bibinfo{person}{Alexander~C Berg}.} \bibinfo{year}{2017}\natexlab{}.
\newblock \showarticletitle{Transformation-grounded image generation network
  for novel 3d view synthesis}. In \bibinfo{booktitle}{\emph{CVPR}}.
\newblock


\bibitem[\protect\citeauthoryear{Pirsiavash and Ramanan}{Pirsiavash and
  Ramanan}{2012}]%
        {Pirsi}
\bibfield{author}{\bibinfo{person}{Hamed Pirsiavash} {and}
  \bibinfo{person}{Deva Ramanan}.} \bibinfo{year}{2012}\natexlab{}.
\newblock \showarticletitle{Detecting activities of daily living in
  first-person camera views}. In \bibinfo{booktitle}{\emph{CVPR}}.
\newblock


\bibitem[\protect\citeauthoryear{Poleg, Arora, and Peleg}{Poleg
  et~al\mbox{.}}{2014}]%
        {Poleg14}
\bibfield{author}{\bibinfo{person}{Yair Poleg}, \bibinfo{person}{Chetan Arora},
  {and} \bibinfo{person}{Shmuel Peleg}.} \bibinfo{year}{2014}\natexlab{}.
\newblock \showarticletitle{Temporal segmentation of egocentric videos}. In
  \bibinfo{booktitle}{\emph{CVPR}}.
\newblock


\bibitem[\protect\citeauthoryear{Regmi and Borji}{Regmi and Borji}{2018}]%
        {Regmi_2018_CVPR}
\bibfield{author}{\bibinfo{person}{Krishna Regmi} {and} \bibinfo{person}{Ali
  Borji}.} \bibinfo{year}{2018}\natexlab{}.
\newblock \showarticletitle{Cross-view image synthesis using conditional gans}.
  In \bibinfo{booktitle}{\emph{CVPR}}.
\newblock


\bibitem[\protect\citeauthoryear{Regmi and Borji}{Regmi and Borji}{2019}]%
        {REGMI2019}
\bibfield{author}{\bibinfo{person}{Krishna Regmi} {and} \bibinfo{person}{Ali
  Borji}.} \bibinfo{year}{2019}\natexlab{}.
\newblock \showarticletitle{Cross-view image synthesis using geometry-guided
  conditional gans}.
\newblock \bibinfo{journal}{\emph{Elsevier CVIU}}  \bibinfo{volume}{187}
  (\bibinfo{year}{2019}), \bibinfo{pages}{102788}.
\newblock


\bibitem[\protect\citeauthoryear{Ronneberger, Fischer, and Brox}{Ronneberger
  et~al\mbox{.}}{2015}]%
        {Unet}
\bibfield{author}{\bibinfo{person}{Olaf Ronneberger}, \bibinfo{person}{Philipp
  Fischer}, {and} \bibinfo{person}{Thomas Brox}.}
  \bibinfo{year}{2015}\natexlab{}.
\newblock \showarticletitle{U-net: Convolutional networks for biomedical image
  segmentation}. In \bibinfo{booktitle}{\emph{International Conference on
  Medical image computing and computer-assisted intervention}}.
\newblock


\bibitem[\protect\citeauthoryear{Ryoo and Matthies}{Ryoo and Matthies}{2013}]%
        {Ryoo13}
\bibfield{author}{\bibinfo{person}{Michael~S Ryoo} {and} \bibinfo{person}{Larry
  Matthies}.} \bibinfo{year}{2013}\natexlab{}.
\newblock \showarticletitle{First-person activity recognition: What are they
  doing to me?}. In \bibinfo{booktitle}{\emph{CVPR}}.
\newblock


\bibitem[\protect\citeauthoryear{Saito, Matsumoto, and Saito}{Saito
  et~al\mbox{.}}{2017}]%
        {TGAN2017}
\bibfield{author}{\bibinfo{person}{Masaki Saito}, \bibinfo{person}{Eiichi
  Matsumoto}, {and} \bibinfo{person}{Shunta Saito}.}
  \bibinfo{year}{2017}\natexlab{}.
\newblock \showarticletitle{Temporal generative adversarial nets with singular
  value clipping}. In \bibinfo{booktitle}{\emph{ICCV}}.
\newblock


\bibitem[\protect\citeauthoryear{Shaham, Dekel, and Michaeli}{Shaham
  et~al\mbox{.}}{2019}]%
        {shaham2019singan}
\bibfield{author}{\bibinfo{person}{Tamar~Rott Shaham}, \bibinfo{person}{Tali
  Dekel}, {and} \bibinfo{person}{Tomer Michaeli}.}
  \bibinfo{year}{2019}\natexlab{}.
\newblock \showarticletitle{Singan: Learning a generative model from a single
  natural image}. In \bibinfo{booktitle}{\emph{ICCV}}.
\newblock


\bibitem[\protect\citeauthoryear{Shama, Mechrez, Shoshan, and
  Zelnik-Manor}{Shama et~al\mbox{.}}{2019}]%
        {shama2019adversarial}
\bibfield{author}{\bibinfo{person}{Firas Shama}, \bibinfo{person}{Roey
  Mechrez}, \bibinfo{person}{Alon Shoshan}, {and} \bibinfo{person}{Lihi
  Zelnik-Manor}.} \bibinfo{year}{2019}\natexlab{}.
\newblock \showarticletitle{Adversarial feedback loop}. In
  \bibinfo{booktitle}{\emph{ICCV}}.
\newblock


\bibitem[\protect\citeauthoryear{Tang, Bai, and Sebe}{Tang
  et~al\mbox{.}}{2020a}]%
        {tang2020dual}
\bibfield{author}{\bibinfo{person}{Hao Tang}, \bibinfo{person}{Song Bai}, {and}
  \bibinfo{person}{Nicu Sebe}.} \bibinfo{year}{2020}\natexlab{a}.
\newblock \showarticletitle{Dual attention gans for semantic image synthesis}.
  In \bibinfo{booktitle}{\emph{ACM MM}}.
\newblock


\bibitem[\protect\citeauthoryear{Tang, Bai, Torr, and Sebe}{Tang
  et~al\mbox{.}}{2020b}]%
        {tang2020bipartite}
\bibfield{author}{\bibinfo{person}{Hao Tang}, \bibinfo{person}{Song Bai},
  \bibinfo{person}{Philip~HS Torr}, {and} \bibinfo{person}{Nicu Sebe}.}
  \bibinfo{year}{2020}\natexlab{b}.
\newblock \showarticletitle{Bipartite graph reasoning gans for person image
  generation}. In \bibinfo{booktitle}{\emph{BMVC}}.
\newblock


\bibitem[\protect\citeauthoryear{Tang, Bai, Zhang, Torr, and Sebe}{Tang
  et~al\mbox{.}}{2020c}]%
        {tang2020xinggan}
\bibfield{author}{\bibinfo{person}{Hao Tang}, \bibinfo{person}{Song Bai},
  \bibinfo{person}{Li Zhang}, \bibinfo{person}{Philip~HS Torr}, {and}
  \bibinfo{person}{Nicu Sebe}.} \bibinfo{year}{2020}\natexlab{c}.
\newblock \showarticletitle{Xinggan for person image generation}. In
  \bibinfo{booktitle}{\emph{ECCV}}.
\newblock


\bibitem[\protect\citeauthoryear{Tang, Liu, and Sebe}{Tang
  et~al\mbox{.}}{2020d}]%
        {tang2020unified}
\bibfield{author}{\bibinfo{person}{Hao Tang}, \bibinfo{person}{Hong Liu}, {and}
  \bibinfo{person}{Nicu Sebe}.} \bibinfo{year}{2020}\natexlab{d}.
\newblock \showarticletitle{Unified generative adversarial networks for
  controllable image-to-image translation}.
\newblock \bibinfo{journal}{\emph{IEEE TIP}}  \bibinfo{volume}{29}
  (\bibinfo{year}{2020}), \bibinfo{pages}{8916--8929}.
\newblock


\bibitem[\protect\citeauthoryear{Tang, Wang, Xu, Yan, and Sebe}{Tang
  et~al\mbox{.}}{2018}]%
        {tang2018gesturegan}
\bibfield{author}{\bibinfo{person}{Hao Tang}, \bibinfo{person}{Wei Wang},
  \bibinfo{person}{Dan Xu}, \bibinfo{person}{Yan Yan}, {and}
  \bibinfo{person}{Nicu Sebe}.} \bibinfo{year}{2018}\natexlab{}.
\newblock \showarticletitle{Gesturegan for hand gesture-to-gesture translation
  in the wild}. In \bibinfo{booktitle}{\emph{ACM MM}}.
\newblock


\bibitem[\protect\citeauthoryear{Tang, Xu, Liu, Wang, Sebe, and Yan}{Tang
  et~al\mbox{.}}{2019a}]%
        {tang2019cycle}
\bibfield{author}{\bibinfo{person}{Hao Tang}, \bibinfo{person}{Dan Xu},
  \bibinfo{person}{Gaowen Liu}, \bibinfo{person}{Wei Wang},
  \bibinfo{person}{Nicu Sebe}, {and} \bibinfo{person}{Yan Yan}.}
  \bibinfo{year}{2019}\natexlab{a}.
\newblock \showarticletitle{Cycle in cycle generative adversarial networks for
  keypoint-guided image generation}. In \bibinfo{booktitle}{\emph{ACM MM}}.
\newblock


\bibitem[\protect\citeauthoryear{Tang, Xu, Sebe, Wang, Corso, and Yan}{Tang
  et~al\mbox{.}}{2019b}]%
        {tang2019multichannel}
\bibfield{author}{\bibinfo{person}{Hao Tang}, \bibinfo{person}{Dan Xu},
  \bibinfo{person}{Nicu Sebe}, \bibinfo{person}{Yanzhi Wang},
  \bibinfo{person}{Jason~J Corso}, {and} \bibinfo{person}{Yan Yan}.}
  \bibinfo{year}{2019}\natexlab{b}.
\newblock \showarticletitle{Multi-channel attention selection gan with cascaded
  semantic guidance for cross-view image translation}. In
  \bibinfo{booktitle}{\emph{CVPR}}.
\newblock


\bibitem[\protect\citeauthoryear{Tang, Xu, Yan, Torr, and Sebe}{Tang
  et~al\mbox{.}}{2020e}]%
        {tang2020local}
\bibfield{author}{\bibinfo{person}{Hao Tang}, \bibinfo{person}{Dan Xu},
  \bibinfo{person}{Yan Yan}, \bibinfo{person}{Philip~HS Torr}, {and}
  \bibinfo{person}{Nicu Sebe}.} \bibinfo{year}{2020}\natexlab{e}.
\newblock \showarticletitle{Local class-specific and global image-level
  generative adversarial networks for semantic-guided scene generation}. In
  \bibinfo{booktitle}{\emph{CVPR}}.
\newblock


\bibitem[\protect\citeauthoryear{Taralova, De~la Torre, and Hebert}{Taralova
  et~al\mbox{.}}{2011}]%
        {taralova2011source}
\bibfield{author}{\bibinfo{person}{Ekaterina Taralova},
  \bibinfo{person}{Fernando De~la Torre}, {and} \bibinfo{person}{Martial
  Hebert}.} \bibinfo{year}{2011}\natexlab{}.
\newblock \showarticletitle{Source constrained clustering}. In
  \bibinfo{booktitle}{\emph{ICCV}}.
\newblock


\bibitem[\protect\citeauthoryear{Tatarchenko, Dosovitskiy, and
  Brox}{Tatarchenko et~al\mbox{.}}{2016}]%
        {TDB16a}
\bibfield{author}{\bibinfo{person}{Maxim Tatarchenko}, \bibinfo{person}{Alexey
  Dosovitskiy}, {and} \bibinfo{person}{Thomas Brox}.}
  \bibinfo{year}{2016}\natexlab{}.
\newblock \showarticletitle{Multi-view 3d models from single images with a
  convolutional network}. In \bibinfo{booktitle}{\emph{ECCV}}.
\newblock


\bibitem[\protect\citeauthoryear{Tulyakov, Liu, Yang, and Kautz}{Tulyakov
  et~al\mbox{.}}{2018}]%
        {MoCoGAN}
\bibfield{author}{\bibinfo{person}{Sergey Tulyakov}, \bibinfo{person}{Ming-Yu
  Liu}, \bibinfo{person}{Xiaodong Yang}, {and} \bibinfo{person}{Jan Kautz}.}
  \bibinfo{year}{2018}\natexlab{}.
\newblock \showarticletitle{Mocogan: Decomposing motion and content for video
  generation}. In \bibinfo{booktitle}{\emph{CVPR}}.
\newblock


\bibitem[\protect\citeauthoryear{Wang, Liu, Zhu, Liu, Tao, Kautz, and
  Catanzaro}{Wang et~al\mbox{.}}{2018}]%
        {wang2018vid2vid}
\bibfield{author}{\bibinfo{person}{Ting-Chun Wang}, \bibinfo{person}{Ming-Yu
  Liu}, \bibinfo{person}{Jun-Yan Zhu}, \bibinfo{person}{Guilin Liu},
  \bibinfo{person}{Andrew Tao}, \bibinfo{person}{Jan Kautz}, {and}
  \bibinfo{person}{Bryan Catanzaro}.} \bibinfo{year}{2018}\natexlab{}.
\newblock \showarticletitle{Video-to-video synthesis}. In
  \bibinfo{booktitle}{\emph{NeurIPS}}.
\newblock


\bibitem[\protect\citeauthoryear{Wang, Jabri, and Efros}{Wang
  et~al\mbox{.}}{2019}]%
        {CVPR2019_CycleTime}
\bibfield{author}{\bibinfo{person}{Xiaolong Wang}, \bibinfo{person}{Allan
  Jabri}, {and} \bibinfo{person}{Alexei~A Efros}.}
  \bibinfo{year}{2019}\natexlab{}.
\newblock \showarticletitle{Learning correspondence from the cycle-consistency
  of time}. In \bibinfo{booktitle}{\emph{CVPR}}.
\newblock


\bibitem[\protect\citeauthoryear{Wang, Bovik, Sheikh, and Simoncelli}{Wang
  et~al\mbox{.}}{2004}]%
        {Wang03}
\bibfield{author}{\bibinfo{person}{Zhou Wang}, \bibinfo{person}{Alan~C Bovik},
  \bibinfo{person}{Hamid~R Sheikh}, {and} \bibinfo{person}{Eero~P Simoncelli}.}
  \bibinfo{year}{2004}\natexlab{}.
\newblock \showarticletitle{Image quality assessment: from error visibility to
  structural similarity}.
\newblock \bibinfo{journal}{\emph{IEEE TIP}} \bibinfo{volume}{13},
  \bibinfo{number}{4} (\bibinfo{year}{2004}), \bibinfo{pages}{600--612}.
\newblock


\bibitem[\protect\citeauthoryear{Yan, Ricci, Liu, and Sebe}{Yan
  et~al\mbox{.}}{2015}]%
        {yanego_2015}
\bibfield{author}{\bibinfo{person}{Yan Yan}, \bibinfo{person}{Elisa Ricci},
  \bibinfo{person}{Gaowen Liu}, {and} \bibinfo{person}{Nicu Sebe}.}
  \bibinfo{year}{2015}\natexlab{}.
\newblock \showarticletitle{Egocentric daily activity recognition via multitask
  clustering}.
\newblock \bibinfo{journal}{\emph{IEEE TIP}} \bibinfo{volume}{24},
  \bibinfo{number}{10} (\bibinfo{year}{2015}), \bibinfo{pages}{2984--2995}.
\newblock


\bibitem[\protect\citeauthoryear{Yonetani, Kitani, and Sato}{Yonetani
  et~al\mbox{.}}{2016}]%
        {Yonetani16}
\bibfield{author}{\bibinfo{person}{Ryo Yonetani}, \bibinfo{person}{Kris~M
  Kitani}, {and} \bibinfo{person}{Yoichi Sato}.}
  \bibinfo{year}{2016}\natexlab{}.
\newblock \showarticletitle{Visual motif discovery via first-person vision}. In
  \bibinfo{booktitle}{\emph{ECCV}}.
\newblock


\bibitem[\protect\citeauthoryear{Zhang, Goodfellow, Metaxas, and Odena}{Zhang
  et~al\mbox{.}}{2019}]%
        {zhang2018self}
\bibfield{author}{\bibinfo{person}{Han Zhang}, \bibinfo{person}{Ian
  Goodfellow}, \bibinfo{person}{Dimitris Metaxas}, {and}
  \bibinfo{person}{Augustus Odena}.} \bibinfo{year}{2019}\natexlab{}.
\newblock \showarticletitle{Self-attention generative adversarial networks}. In
  \bibinfo{booktitle}{\emph{ICML}}.
\newblock


\bibitem[\protect\citeauthoryear{Zhou, Tulsiani, Sun, Malik, and Efros}{Zhou
  et~al\mbox{.}}{2016}]%
        {zhou2016view}
\bibfield{author}{\bibinfo{person}{Tinghui Zhou}, \bibinfo{person}{Shubham
  Tulsiani}, \bibinfo{person}{Weilun Sun}, \bibinfo{person}{Jitendra Malik},
  {and} \bibinfo{person}{Alexei~A Efros}.} \bibinfo{year}{2016}\natexlab{}.
\newblock \showarticletitle{View synthesis by appearance flow}. In
  \bibinfo{booktitle}{\emph{ECCV}}.
\newblock


\bibitem[\protect\citeauthoryear{Zhu, Park, Isola, and Efros}{Zhu
  et~al\mbox{.}}{2017}]%
        {CycleGAN2017}
\bibfield{author}{\bibinfo{person}{Jun-Yan Zhu}, \bibinfo{person}{Taesung
  Park}, \bibinfo{person}{Phillip Isola}, {and} \bibinfo{person}{Alexei~A
  Efros}.} \bibinfo{year}{2017}\natexlab{}.
\newblock \showarticletitle{Unpaired image-to-image translation using
  cycle-consistent adversarial networks}. In \bibinfo{booktitle}{\emph{ICCV}}.
\newblock


\end{thebibliography}










\end{document}